\documentclass{article} %
\usepackage[OT1]{fontenc}
\usepackage{iclr2027_conference,times}

\usepackage{amsmath,amsfonts,bm}

\def\eqref#1{equation~\ref{#1}}

\def\1{\bm{1}}

\DeclareMathAlphabet{\mathsfit}{\encodingdefault}{\sfdefault}{m}{sl}
\SetMathAlphabet{\mathsfit}{bold}{\encodingdefault}{\sfdefault}{bx}{n}

\usepackage{hyperref}
\usepackage{url}
\usepackage{xspace}
\usepackage{enumitem}
\usepackage{graphicx}
\usepackage{wrapfig}
\usepackage[section]{placeins}
\usepackage{subcaption}
\usepackage{booktabs}
\usepackage{array}
\usepackage{xcolor}
\usepackage{pifont}
\usepackage{amsmath}
\usepackage{amssymb}
\usepackage{microtype}
\usepackage{multirow}
\usepackage{wrapfig}
\usepackage{acronym}
\newcommand{\method}[0]{\textsc{MolPAIR}\xspace}
\newcommand{\methodlong}[0]{Molecular Pair-Augmented In-context Refinement\xspace}

\newcommand{\mypar}[1]{\textbf{#1}\hspace{4pt}}

\acrodef{TFM}[TFM]{tabular foundation model}
\acrodef{GNN}[GNN]{graph neural network}
\acrodef{MLP}[MLP]{multilayer perceptron}
\acrodef{MMP}[MMP]{matched molecular pair}
\acrodef{OOF}[OOF]{out-of-fold}
\acrodef{HPO}[HPO]{hyperparameter optimization}
\acrodef{PCA}[PCA]{principal component analysis}
\acrodef{ADME}[ADME]{absorption, distribution, metabolism, and excretion}
\acrodef{RMSE}[RMSE]{root mean squared error}
\acrodef{MAE}[MAE]{mean absolute error}
\acrodef{MSE}[MSE]{mean squared error}
\acrodef{PR-AUC}[PR-AUC]{area under the precision-recall curve}
\acrodef{ROC-AUC}[ROC-AUC]{area under the receiver operating characteristic curve}
\acrodef{CI}[CI]{confidence interval}
\acrodef{ECFP4}[ECFP4]{extended-connectivity fingerprint of diameter four}
\acrodef{MACCS}[MACCS]{Molecular ACCess System}
\acrodef{GIN}[GIN]{graph isomorphism network}
\acrodef{GINE}[GINE]{graph isomorphism network with edge features}
\acrodef{SMILES}[SMILES]{simplified molecular-input line-entry system}
\acrodef{ERM}[ERM]{empirical risk minimization}
\acrodef{GPU}[GPU]{graphics processing unit}
\acrodef{CPU}[CPU]{central processing unit}
\acrodef{NUMA}[NUMA]{non-uniform memory access}

\definecolor{molrift}{RGB}{190,74,43}

\title{Molecular Property Prediction under\\Structural Shift with\\Tabular Foundation Models}

\author{
Jinmo Lee$^{1}$\thanks{These authors contributed equally.}\quad
Dooho Lee$^{1}$\footnotemark[1]\quad
Minho Jeong$^{1}$\footnotemark[1]\quad
Jaemin Yoo$^{1}$\\
$^{1}$Nums AI\\
\texttt{\{jinmo.lee, dooho, minho.jeong, jaemin\}@nums.world}
}

\iclrfinalcopy

\begin{document}
\maketitle

\begin{abstract}
Predicting molecular properties for compounds that differ structurally from labeled training molecules is important for drug discovery and materials design.
Tabular foundation models (TFMs) offer a promising approach through in-context learning, but their performance under structural shifts and the value of molecular comparisons in this setting remain underexplored.
We study structural generalization in molecular property prediction and introduce \method{} (\methodlong{}), a framework that combines molecule-level and molecular-pair contexts without task-specific parameter updates.
A global \ac{TFM} first predicts a query's property from labeled molecular examples.
A second frozen \ac{TFM} predicts differences in prediction errors between the query and labeled reference molecules, using these comparisons to refine the initial prediction.
Across 58 MoleculeACE and Polaris tasks, CheMeleon representations combined with TabPFN-3 already outperform each evaluated baseline on a majority of tasks.
\method{} further improves this predictor on 46 of 58 tasks, with gains across four molecular representations and three TFM backbones.
These results show that explicit molecular comparisons can strengthen tabular in-context learning for structural generalization while keeping the molecular encoder and pretrained model weights fixed.
The code and datasets are available at \url{https://github.com/nums-ai/MolPAIR}.
\end{abstract}

\acresetall

\section{Introduction}
\label{sec:introduction}

\begin{wrapfigure} {r}{0.5\textwidth}
    \vspace{-4.5em}
    \centering
    \includegraphics[
        width=0.96\linewidth
    ]{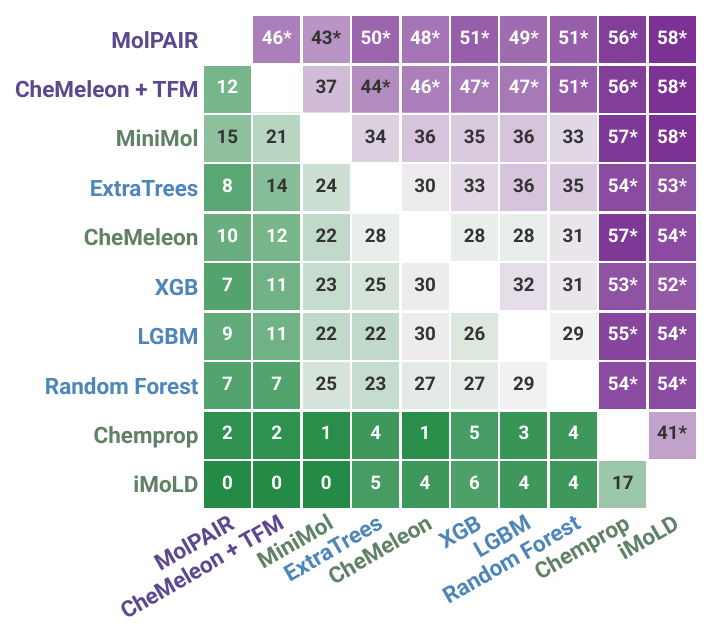}
    \setlength{\abovecaptionskip}{3pt}
    \caption{\textbf{Pairwise task-win matrix.}
    Each cell counts the tasks on which the row method outperforms the column method based on seed-averaged primary metrics.
    * indicates significance (two-sided exact sign test, Holm-adjusted $p<0.05$).}
    \label{fig:pair-correction-gain}
    \vspace{-1.2em}
\end{wrapfigure}
Molecular property prediction helps researchers decide which molecules to test in the laboratory by estimating properties from chemical structure~\citep{wu2018moleculenet}.
Most approaches use labeled molecules to train a model for each prediction task.
Chemprop, for example, learns molecular representations and a predictor together through end-to-end training~\citep{yang2019analyzing}.
Pretrained molecule encoders such as MiniMol and CheMeleon instead provide reusable representations for a fitted prediction head and can also be fine-tuned for the target task~\citep{klaser2404minimol,burns2025deep}.
In both cases, adapting to a new prediction task requires task-specific training and hyperparameter tuning.

Recent studies use \acp{TFM} to reduce this per-dataset optimization burden through in-context learning.
\Acp{TFM} are pretrained on diverse synthetic tables to infer relationships between features and labels from examples supplied as context~\citep{hollmann2022tabpfn}.
For a new task, they use labeled rows as context to predict query labels without updating their pretrained weights~\citep{hollmann2025accurate,qu2025tabicl}.
In molecular applications, each row represents one molecule, with its representation as the features and its measured property as the label.
Combining molecular representations with \acp{TFM} has produced competitive results across benchmarks~\citep{hicham2026tabular,guan2026can,banaszewski2026monroe}.

When solving drug discovery or materials design through molecular property prediction, researchers need to explore new chemical series because known compounds do not always offer the desired combination of properties~\citep{schneider2006scaffold,fralish2023deepdelta}.
This requires predicting properties for molecules from chemical series absent from the labeled data, a setting we refer to as \emph{structural generalization}.
One approach is to compare new and known molecules.
Pairwise methods such as PADRE and DeepDelta follow this idea by learning how differences in molecular structure relate to differences in their properties~\citep{tynes2021pairwise,fralish2023deepdelta}.
However, it remains unclear how effectively \acp{TFM} handle such structural shifts and how these molecular comparisons can be incorporated into tabular in-context learning.

To address this gap, we introduce \method{} (\methodlong{}), a framework that incorporates molecular comparisons into tabular in-context learning to improve structural generalization in molecular property prediction.
One \ac{TFM}, the global predictor, makes an initial prediction based on the global context, in which each row represents one molecule.
We then obtain held-out predictions for the labeled molecules and construct a relation context with each row representing a molecular pair and its prediction-error difference.
A second \ac{TFM}, the relation predictor, uses these pairwise examples to estimate how the prediction error changes from a reference molecule to a query molecule.
Combining this difference with the reference molecule's known error yields the query's estimated error, which is used to correct the initial prediction.

We evaluate \method{} on 30 MoleculeACE and 28 Polaris tasks using scaffold-separated, similarity-filtered splits~\citep{van2022exposing,polaris2026}.
These splits operationalize structural generalization by separating training and test molecules according to their structures.
Our primary configuration uses CheMeleon representations with TabPFN-3 as the TFM backbone and RDKit2D features for the relation context~\citep{burns2025deep,grinsztajn2026tabpfn,rdkit2024}.
Figure~\ref{fig:pair-correction-gain} shows that combining molecular representations with a frozen \ac{TFM} already outperforms each evaluated baseline on a majority of tasks.
Building on this strong baseline, \method{} further improves predictions on 46 of 58 tasks, showing that molecular comparisons can further improve a TFM predictor under structural shift without changing its representation or pretrained parameters.
\section{Related Work}
\label{sec:related}

\mypar{Tabular foundation models.}
Tabular foundation models (TFMs) such as TabPFN build on Prior-Data Fitted Networks, which learn to approximate Bayesian prediction under a prior over tasks~\citep{muller2021transformers,hollmann2022tabpfn}.
Given a labeled context $\mathcal{D}$ and a query $x_q$, the prediction target is the posterior predictive distribution
\begin{equation}
p(y_q \mid x_q,\mathcal{D})
=
\int p(y_q \mid x_q,t)\,p(t\mid\mathcal{D})\,dt,
\label{eq:pfn-ppd}
\end{equation}
where $t$ denotes a latent prediction task.
TabPFN is pretrained to approximate this distribution using synthetic tables generated from structural causal models~\citep{hollmann2025accurate}.
At inference time, labeled rows provide context for a new task, allowing the model to predict query labels without updating its pretrained weights.
More recent \acp{TFM}, including TabICLv2 and Causilo, use scalable architectures to support efficient inference on larger tables~\citep{qu2026tabiclv2,cho2026causilotechnicalreport}.

\mypar{Molecular property predictors.}
Classical predictors combine molecular fingerprints or descriptors with forests and boosting, while \acp{GNN} exploit graph topology~\citep{rogers2010extended,breiman2001random,chen2016xgboost,yang2019analyzing}.
Pretrained encoders learn reusable embeddings from large molecular datasets through diverse training schemes.
For example, MolCLR uses graph contrastive learning, MoLFormer learns from molecular strings, MiniMol uses sparse multitask labels, and CheMeleon predicts chemical descriptors~\citep{wang2022molecular,ross2022large,klaser2404minimol,burns2025deep}.
Recent studies apply \acp{TFM} using molecular representations as inputs $x_i$ and measured properties as labels $y_i$~\citep{hicham2026tabular}.
Some of these studies include analyses on scaffold splits and out-of-distribution evaluations~\citep{guan2026can,banaszewski2026monroe}, but their primary focus is on molecular representations and predictors.
Therefore, how molecular comparisons can be incorporated into tabular in-context learning, and whether they improve structural generalization, remain open questions.

\mypar{Structural generalization and evaluation.}
Evaluating structural generalization requires specifying how test molecules differ from the labeled molecules.
Random splits can place close analogues in both sets, so performance on unseen molecules alone does not establish generalization to unfamiliar structures~\citep{wallach2018most,wu2018moleculenet}.
The split therefore defines the generalization scenario being tested.
Scaffold splits hold out groups with a shared molecular framework, while chronological splits use earlier measurements to predict later compounds~\citep{bemis1996properties,sheridan2013time}.
Lo-Hi distinguishes hit identification from lead optimization, SIMPD approximates temporal changes, and DataSAIL reduces similarity leakage between partitions~\citep{steshin2023hi,landrum2023simpd,joeres2025data}.
Our evaluation combines scaffold splitting with a threshold on test-to-training fingerprint similarity to exclude both shared scaffolds and similar test molecules.

\mypar{Learning from molecular comparisons.}
Beyond defining how structural generalization is evaluated, a separate question is how information from known molecules can be used to predict unfamiliar ones.
One approach is to compare molecules and predict relative rather than absolute quantities.
\Acp{MMP} identify compounds related by a local transformation~\citep{hussain2010computationally}.
PADRE formulates regression through differences between pairs of examples, while DeepDelta learns molecular property differences directly~\citep{tynes2021pairwise,fralish2023deepdelta}.
We build on this idea of molecular comparison by using pairwise information as context for a \ac{TFM}.
\section{\method{}}
\label{sec:method}
\begin{figure}[t]
    \centering
    \includegraphics[width=\textwidth]{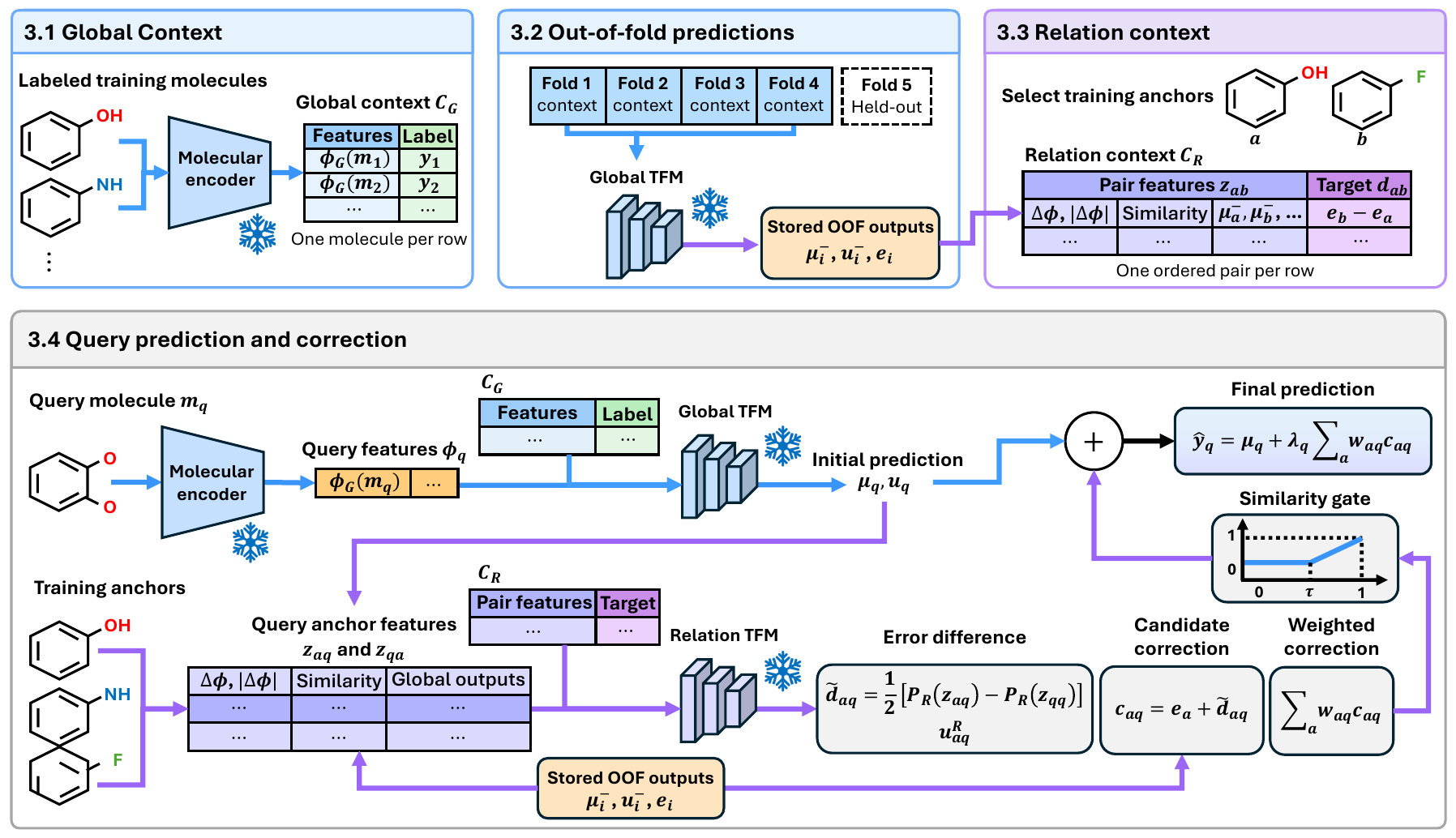}
    \caption{
    \textbf{\method{} framework.}
    The blue path predicts the query's property while the purple path uses the relation context to predict error differences.
    Snowflakes mark frozen components.
    }
    \label{fig:method}
\end{figure}

\method{} uses two \acp{TFM} with different context tables, as shown in Figure~\ref{fig:method}.
The global predictor reads a one-molecule-per-row table to make an initial prediction for a query.
We then pair each training molecule with labeled reference molecules, called \emph{anchors}, to construct the relation context.
The relation predictor uses this context to predict the error difference between a query and its anchor.
We estimate the query's error and correct its prediction by adding this difference to the anchor's error.

\subsection{Constructing the global context}

Let $\mathcal{D}=\{(m_i,y_i)\}_{i=1}^{n}$ denote the labeled training molecules for a regression or binary classification task.
A fixed encoder $\phi_G$ maps each molecule $m_i$ to a feature vector and forms the global context
\begin{equation}
\mathcal{C}_G=\{(\phi_G(m_i),y_i)\}_{i=1}^{n}.
\end{equation}
Given this global context and the encoded features $\phi_G(m_q)$ of a query molecule $m_q$, the global TFM predictor $P_G$ returns a point prediction $\mu_q$ and an uncertainty summary $u_q$.
For regression, $\mu_q$ is the predictive mean, and
$u_q=(Q_{0.9}(m_q)-Q_{0.1}(m_q))/2.563$,
where $Q_{\alpha}(m_q)$ denotes the $\alpha$-quantile of the predictive distribution and the factor $2.563$ converts the 10--90\% quantile width to a standard-deviation scale under a normal distribution.
For binary classification, $\mu_q$ is the positive-class probability, and $u_q$ is the entropy of the predicted class distribution.

\subsection{Obtaining out-of-fold predictions}

Before constructing the relation context, we estimate the global predictor's errors on the training molecules.
These errors serve two purposes:
(1) their differences provide targets for the relation context, and
(2) the individual errors later serve as references for correcting query predictions.

To obtain an error for each training molecule without using its own label as context, we divide the global context into five folds and predict each fold using the other four as context.
For molecule $i$ in fold $f(i)$, we compute
\begin{equation}
(\mu_i^{-},u_i^{-})
=
P_G^{-f(i)}(\phi_G(m_i)),
\qquad
e_i=y_i-\mu_i^{-}.
\label{eq:oof}
\end{equation}
Here, $P_G^{-f(i)}$ denotes the same global predictor with fold $f(i)$ excluded from its context.
The superscript $-$ denotes these \ac{OOF} outputs.
The error $e_i$ is positive when the prediction is too low and negative when it is too high.
For binary classification, $e_i$ is the binary label minus the predicted positive-class probability.
We store $\mu_i^{-}$, $u_i^{-}$, and $e_i$ for each training molecule.

\subsection{Constructing the relation context}

Using the stored OOF outputs, we construct a table with one pair of training molecules per row.
Each input describes the two molecules and the global predictor's outputs for them, while the target is the difference between their prediction errors.

\mypar{Selecting anchors.}
An \textit{anchor} is a labeled molecule used as a reference for another molecule.
For each training molecule $b$, we select up to eight other training molecules as candidate anchors from the same OOF fold.
We prioritize anchors that form an exact one-cut \ac{MMP} with $b$, followed by other molecules with high Morgan fingerprint similarity~\citep{rogers2010extended,hussain2010computationally}.
This procedure selects exact matched pairs when available and chemically similar pairs otherwise.

\mypar{Defining the pair target.}
For each training molecule $b$ and one of its selected anchors $a$, we form an ordered pair $(a,b)$.
The pair target is
\begin{equation}
d_{ab}=e_b-e_a.
\end{equation}
Thus, $d_{ab}$ describes how the global predictor's signed error changes from the anchor $a$ to molecule $b$.
Since $e_b=e_a+d_{ab}$, adding the predicted error difference to the anchor's known error provides an estimate of $b$'s error.
At inference time, we apply the same operation to estimate the error of a query molecule.
The relation task remains a regression problem for binary endpoints because the error differences are continuous.

\mypar{Describing molecular pairs.}
A fixed feature map $\phi_R$ represents both molecules in a pair.
This map may differ from the global encoder $\phi_G$.
We define the signed feature difference as
$\Delta\phi_{ab}=\phi_R(m_b)-\phi_R(m_a)$
and construct the relation feature vector
\begin{equation}
\begin{aligned}
z_{ab} = [&
\Delta\phi_{ab},
|\Delta\phi_{ab}|,
s_{ab},
r_{ab}^{\mathrm{scaf}},
r_{ab}^{\mathrm{MMP}},
\mu_a^{-},
\mu_b^{-},
\mu_b^{-}-\mu_a^{-},
u_a^{-},
u_b^{-}
].
\end{aligned}
\label{eq:relation-features}
\end{equation}
These features describe the molecular differences, the chemical relationship, and the global predictions for the pair.
The signed and absolute feature differences capture the direction and magnitude of changes in molecular features.
The Morgan Tanimoto similarity $s_{ab}$ measures molecular similarity, while
$r_{ab}^{\mathrm{scaf}}$ and $r_{ab}^{\mathrm{MMP}}$
indicate a shared Bemis--Murcko scaffold and an exact one-cut \ac{MMP}, respectively~\citep{bemis1996properties}.
The remaining features contain the global predictor's OOF predictions, their difference, and their uncertainty summaries.

Let $\mathcal{P}$ denote the selected ordered training pairs.
Their inputs and targets form the relation context \begin{equation}
    \mathcal{C}_R=\{(z_{ab},d_{ab})\}_{(a,b)\in\mathcal{P}}.
\end{equation}
The resulting table $\mathcal{C}_R$ provides the second frozen \ac{TFM}, called the relation predictor $P_R$, with examples of molecular pairs and their corresponding error differences.

\subsection{Predicting and correcting a new query}

\mypar{Producing anchor-specific corrections.}
Given a query molecule $m_q$, the global predictor reads the full global context and returns an initial prediction $\mu_q$ and uncertainty summary $u_q$.
We then select up to 32 anchors from the full training split using the same chemical ranking.
For each anchor $a$, we construct $z_{aq}$ using Equation~\ref{eq:relation-features}, with the query taking the place of molecule $b$.
The anchor supplies its stored OOF outputs $\mu_a^{-}$ and $u_a^{-}$, while the query supplies $\mu_q$ and $u_q$ from the full global context.

The relation predictor estimates the error difference in both directions.
We obtain $z_{qa}$ by reversing the molecular order in the same feature construction.
Because reversing a pair should reverse the sign of its error difference, we combine the predictions as
\begin{equation}
    \widetilde d_{aq}
    =
    \frac{P_R(z_{aq})-P_R(z_{qa})}{2}.
    \label{eq:symmetry}
\end{equation}
We also average the corresponding uncertainty scales,
\begin{equation}
    u^R_{aq}=
\frac{u^{P_R}(z_{aq})+u^{P_R}(z_{qa})}{2}
\end{equation}
where $u^{P_R}(z)$ denotes the relation predictor's uncertainty scale for input $z$.
The estimate $\widetilde d_{aq}$ describes the query's error relative to the anchor's.
Adding this predicted error difference to the anchor's stored error gives a candidate correction to the query's initial prediction, $c_{aq}=e_a+\widetilde d_{aq}$.

\mypar{Combining and gating corrections.}
We combine the candidate corrections by giving more weight to anchors that are chemically similar to the query and whose relation predictions are less uncertain.
Let $\mathcal{A}_q$ denote the accepted anchors for a query $q$. For each accepted anchor, we compute
\begin{equation}
    \widetilde w_{aq}
    =
    \frac{\max(s_{aq},10^{-4})^2}
         {\max(u^R_{aq},10^{-4})},
    \qquad
    w_{aq}
    =
    \frac{\widetilde w_{aq}}
         {\sum_{a'\in\mathcal{A}_q}\widetilde w_{a'q}}.
    \label{eq:anchor-weights}
\end{equation}

The squared Morgan Tanimoto similarity term favors closer anchors, while the inverse-uncertainty term reduces the influence of less reliable error-difference estimates.

The weights determine how to combine the corrections, but do not determine whether the anchors are sufficiently similar to justify a large correction.
A separate reliability gate therefore controls how strongly the combined correction is applied.
Using the threshold $\tau=0.2$, we compute
\begin{equation}
    \lambda_q
    =
    \operatorname{clip}\!\left(
        \frac{\max_{a\in\mathcal{A}_q}s_{aq}-\tau}
             {1-\tau},
        0,
        1
    \right),
    \qquad
    \widehat y_q
    =
    \mu_q
    +
    \lambda_q
    \sum_{a\in\mathcal{A}_q}
    w_{aq}c_{aq}.
    \label{eq:prediction}
\end{equation}
The gate permits a larger correction as the closest accepted anchor becomes more similar to the query.
If no anchor is accepted or none has similarity above $\tau$, \method{} returns $\mu_q$ unchanged.
\section{Evaluation Protocol}
\label{sec:experiments}

This section describes the datasets, baseline methods, and default model settings used throughout the evaluation.
Experiment-specific configurations are introduced in Sections~\ref{sec:results} and~\ref{sec:ablations}.

\subsection{Datasets and performance measures}

We combine 30 MoleculeACE bioactivity tasks with 28 Polaris tasks~\citep{van2022exposing,polaris2026}, comprising 47 regression and 11 binary classification tasks.
For each task, we retain the released target transformation and primary evaluation metric.
To evaluate structural generalization, we construct splits that separate test molecules structurally from the labeled training data.
We hold out complete Bemis--Murcko scaffold groups from the released training and validation molecules, then exclude test molecules whose maximum Morgan Tanimoto similarity to any training molecule exceeds 0.60.
The scaffold holdout separates structural groups, while the similarity filter removes close analogues that can remain across different scaffolds.
Across five seeds, this procedure yields 286 splits over 58 tasks.
Appendix~\ref{app:primary-splits} provides the construction details and applicability criteria.

We average the available seeds within each task before computing Win/Loss and native-metric summaries.
We report mean \ac{RMSE} across MoleculeACE and retain the individual metric scales for Polaris.
Elo uses a penalized Bradley and Terry model with equal total weight per endpoint~\citep{bradley1952rank,erickson2026tabarena}.
All Elo analyses first average each method's available seed metrics within each endpoint, then assign one direction-aware win, loss, or tie per endpoint and method pair.
Ratings are centered at 1000 within each stated method pool.
We further obtain 95\% Elo intervals by resampling complete endpoints 10,000 times and refitting the model.
Appendices~\ref{app:datasets} and~\ref{app:aggregation} provide the task metrics and aggregation details, respectively.

\subsection{Baseline methods}

Our primary benchmark compares \method{} with its global predictor and eight additional baselines.
The global baseline combines the same molecular embeddings and the \ac{TFM} backbone, but without the relation context and predictor.
The molecular baselines include a fitted prediction head on MiniMol embeddings, CheMeleon fine-tuning, and the end-to-end trained Chemprop model~\citep{klaser2404minimol,burns2025deep,yang2019analyzing}.
We also include Random Forest, ExtraTrees, XGBoost, and LightGBM, together with iMoLD, designed for molecular distribution shifts~\citep{breiman2001random,geurts2006extremely,chen2016xgboost,ke2017lightgbm,zhuang2023learning}.
We evaluate default and train-only tuned configurations where available, while CheMeleon fine-tuning follows its canonical configuration.
Appendix~\ref{app:baseline-settings} gives the baseline training settings and tuning budgets.

\subsection{Default model configuration}

Unless otherwise stated, \method{} uses CheMeleon representations for the global context, RDKit2D descriptors for the relation context, and TabPFN-3 for both \acp{TFM}~\citep{burns2025deep,rdkit2024,grinsztajn2026tabpfn}.
All relation-context pairs are formed within the same fold.
For the robustness experiments, we vary the global molecular representation among RDKit2D, CheMeleon, MiniMol, and Monroe~\citep{klaser2404minimol,banaszewski2026monroe}, and replace TabPFN-3 with TabICLv2 or Causilo~\citep{qu2026tabiclv2,cho2026causilotechnicalreport}.
At query inference, we exclude anchors with Morgan Tanimoto similarity below 0.20 and set the reliability-gate threshold to $\tau=0.20$.
Appendix~\ref{app:implementation-settings} specifies the feature processing, context budgets, and other inference settings.
\section{Results}
\label{sec:results}

To assess the efficacy of tabular in-context learning for structural generalization, we compare the global TFM predictor with other baselines and examine the additional gains from \method{}.
We then test whether these gains persist across definitions of structural separation, representations, and TFMs.

\subsection{How well does \method{} generalize to unfamiliar molecules?}
\label{sec:primary-results}

\begin{table}[t]
\centering
\scriptsize
\setlength{\tabcolsep}{2pt}
\caption{\textbf{Structural generalization across 58 tasks.} MACE RMSE is mean $\pm$ SD across tasks after averaging seeds; ranks are task means and Top one counts first places. Brackets give 95\% Elo intervals; bold marks the best values.}
\label{tab:new-series-benchmark}
\resizebox{\textwidth}{!}{%
\begin{tabular}{@{}lrrrrrr@{}}
\toprule
Method & MACE RMSE $\downarrow$ & MACE rank $\downarrow$ & Polaris cls. rank $\downarrow$ & Polaris reg. rank $\downarrow$ & Top one $\uparrow$ & Elo $\uparrow$ \\
\midrule
\textbf{\method{}} & \textbf{0.885} $\pm$ 0.100 & \textbf{2.00} & \textbf{2.18} & \textbf{2.59} & \textbf{29} & \textbf{1360 [1289, 1455]} \\
Global baseline  & 0.893 $\pm$ 0.097 & 2.93 & 2.36 & 4.00 & 7 & 1246 [1196, 1310] \\
MiniMol (tuned) & 0.931 $\pm$ 0.107 & 5.30 & 3.91 & 3.12 & 14 & 1122 [1060, 1190] \\
ExtraTrees (tuned)  & 0.921 $\pm$ 0.102 & 4.90 & 5.36 & 5.12 & 2 & 1062 [1007, 1123] \\
CheMeleon (fine-tuned) & 0.930 $\pm$ 0.089 & 5.40 & 6.91 & 4.24 & 3 & 1035 [976, 1097] \\
XGBoost (tuned)  & 0.924 $\pm$ 0.101 & 5.03 & 6.36 & 5.59 & 2 & 1026 [969, 1088] \\
LightGBM (tuned)  & 0.927 $\pm$ 0.102 & 5.10 & 6.36 & 5.82 & 1 & 1016 [965, 1069] \\
Random Forest (tuned) & 0.937 $\pm$ 0.096 & 5.67 & 4.09 & 6.59 & 0 & 1008 [957, 1060] \\
Chemprop (tuned)  & 1.046 $\pm$ 0.107 & 9.03 & 9.09 & 8.59 & 0 & 602 [491, 681] \\
iMoLD & 1.080 $\pm$ 0.139 & 9.63 & 8.36 & 9.35 & 0 & 522 [376, 615] \\
\bottomrule
\end{tabular}}
\end{table}

Table~\ref{tab:new-series-benchmark} shows that the global predictor ranks second at 1246 Elo, ahead of all evaluated task-trained molecular models and tree ensembles.
Thus, combining molecular representations with tabular in-context learning already provides a strong baseline for structural generalization, without task-specific parameter updates.
\method{} improves this baseline on 46 of 58 tasks (Figure~\ref{fig:pair-correction-gain}), raising its pooled Elo to 1360 and increasing the number of first-place finishes from seven to 29.
It also achieves the best mean ranks across MoleculeACE, Polaris classification, and Polaris regression, while reducing mean MoleculeACE \ac{RMSE} from 0.893 to 0.885.
The improvement therefore appears in both cross-task rankings and native regression error, rather than only in the aggregate Elo rating.

Importantly, the gain over a strong global predictor suggests that molecule-level prediction does not fully capture the information available in the labeled molecules.
Explicit molecular comparisons provide a complementary signal that remains useful across both regression and classification under structural shift.
Appendix~\ref{app:specialized-results} provides additional comparisons with specialized molecular methods.

\subsection{Is the improvement robust to the definition of structural shift?}
\label{sec:split-robustness-results}
\label{sec:cutoff-results}

\begin{wraptable}{R}{0.45\textwidth}
\vspace{-1em}
\centering
\small
\fontsize{8}{10}\selectfont
\setlength{\tabcolsep}{2pt}

\caption{\textbf{Sensitivity to the training-similarity cutoff.} RMSE is task mean $\pm$ SD on MoleculeACE; W/L counts \method{} wins/losses across retained tasks. Bold marks lower RMSE.}
\label{tab:similarity-threshold-robustness}
\begin{tabular}{@{}lccc@{}}
\toprule
Cutoff & Global RMSE & \method{} RMSE & W/L \\
\midrule
0.40 & 0.9939 $\pm$ 0.1745 & \textbf{0.9926 $\pm$ 0.1792} & 38/14 \\
0.50 & 0.9570 $\pm$ 0.1318 & \textbf{0.9521 $\pm$ 0.1302} & 41/13 \\
0.60 & 0.8932 $\pm$ 0.0966 & \textbf{0.8849 $\pm$ 0.1004} & 46/12 \\
0.70 & 0.8357 $\pm$ 0.0791 & \textbf{0.8327 $\pm$ 0.0790} & 42/16 \\
\bottomrule
\end{tabular}
\end{wraptable}
The dataset split protocol determines which forms of structural separation between training and test molecules are evaluated.
We therefore test whether the gains of \method{} over global predictor persist across alternative split rules.
Our primary split holds out scaffold groups and removes test molecules whose maximum Morgan Tanimoto similarity to the training set exceeds 0.60.
Varying this cutoff from 0.40 to 0.70, \method{} wins a majority of tasks and lowers mean MoleculeACE \ac{RMSE} at every setting (Table~\ref{tab:similarity-threshold-robustness}).
Appendix~\ref{app:alternative-split-protocols} evaluates an alternative split construction, replacing the scaffold-plus-cutoff protocol with DataSAIL C1 partitions defined by ECFP4 or MACCS similarity~\citep{joeres2025data}.
With Morgan-based anchor selection unchanged, \method{} outperforms the global predictor on 38 of 58 tasks under each partitioning scheme.
These results suggest that the improvement is not specific to our primary split setting and does not require matching fingerprints for splitting and anchor selection.

\subsection{Does the improvement persist across representations and TFMs?}
\label{sec:representation-backbone-results}
\label{sec:backbone-results}
\begin{figure}[h]
    \centering
    \includegraphics[width=0.97\textwidth]{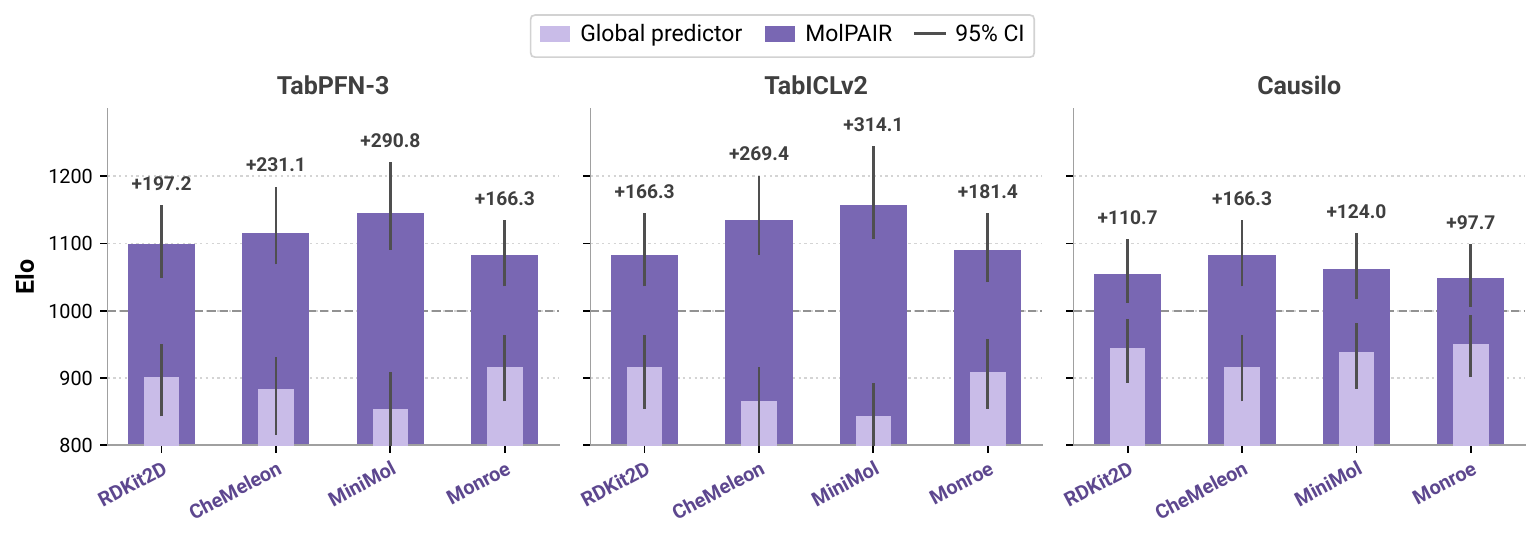}
    \caption{\textbf{Elo across representations and backbones.} Labels give \method{}-minus-global $\Delta$Elo; error bars show 95\% intervals. Absolute ratings are comparable only within pairs.}
    \label{fig:new-series-headline}
\end{figure}

To determine whether the gains extend beyond the primary CheMeleon--TabPFN-3 configuration, we evaluate \method{} across different molecular representations and TFM backbones.
We cross four representations---RDKit2D, CheMeleon, MiniMol, and Monroe---with three backbones---TabPFN-3, TabICLv2, and Causilo.
For each combination, we compare \method{} with its global baseline while keeping the molecular representation and backbone fixed.
All combinations use RDKit2D relation features and same-fold OOF pairing.
For each representation--backbone combination, we fit Elo separately to \method{} and its corresponding global predictor on the 58-task benchmark.

Figure~\ref{fig:new-series-headline} shows positive Elo gains for \method{} in all twelve combinations, ranging from +97.7 to +314.1.
With TabPFN-3, the gains range from +166.3 to +290.8 across the four representations.
The gains remain positive with TabICLv2, where MiniMol yields the largest improvement overall at +314.1, and with Causilo, where they range from +97.7 to +166.3.
Notably, RDKit2D also shows positive gains with all three backbones, indicating that the benefit does not depend on a pretrained molecular embedding.
Thus, while the magnitude of the improvement varies across configurations, the consistent gains suggest that molecular-pair context provides a useful signal across different molecular representations and TFM backbones.
Appendix~\ref{app:representation-backbone-results} provides the full comparison table, including MoleculeACE RMSE for all twelve configurations.

\section{Additional Analyses and Computational Cost}
\label{sec:additional-analyses}
\label{sec:ablations}
\label{app:correction-studies}

\method{} uses predicted error differences to refine the predictions of a global TFM.
However, similar gains might be obtained from other approaches, such as directly using prediction errors observed on similar labeled molecules.
Therefore, we first compare \method{} with these simpler alternatives.
We further shuffle the error-difference targets assigned to molecular pairs to test whether learning the correct pair relationships matters, and also test whether the reliability gate improves predictions.
Last but not least, we analyze the additional computation cost of \method{} when compared to other supervised methods and the global baseline.

To answer these questions, we construct alternatives that use the labeled molecules without learning pairwise error differences, and variants that change the pair targets or remove the gate.
Alongside \method{} and the global baseline used in Section~\ref{sec:results}, we evaluate five additional configurations: 
\begin{enumerate}[label=(\arabic*), leftmargin=*, labelindent=0pt, nosep]
    \item \textbf{Two-global-TFM ensemble} averages predictions from two independently seeded global TFMs without molecular comparisons.
    \item \textbf{$k$NN residual} adjusts the global prediction using a weighted average of stored prediction errors from similar labeled molecules, without learning pairwise error differences.
    \item \textbf{Single-molecule residual} uses a second TFM to predict the query's error directly from molecule-level features.
    \item \textbf{Shuffled relation labels} randomly reassigns error-difference targets among molecular pairs, breaking the correspondence between each pair and its target.
    \item \textbf{No reliability gate} retains the pair predictions and anchor weighting but applies the resulting adjustment without reducing its strength when anchors are dissimilar to the query.
\end{enumerate}

\method{}, the global baseline, and all five alternatives use the same structural-generalization splits across 58 tasks.
Implementation details are  in Appendix~\ref{app:shared-global-protocol}.

\subsection{Do molecular comparisons outperform simpler alternatives?}
\label{sec:single-correction-results}
\begin{table}[t]
\centering
\scriptsize
\fontsize{8}{9.5}\selectfont
\setlength{\tabcolsep}{1pt}
\caption{\textbf{Prediction and correction controls.}
Pooled $\Delta$Elo and W/L compare full \method{} with each control on 58 tasks; RMSE reductions use MoleculeACE.
Positive values favor \method{}; brackets give 95\% endpoint-bootstrap intervals.}
\label{tab:design-studies-main}
\begin{tabular*}{\textwidth}{@{\extracolsep{\fill}}lrrr@{}}
\toprule
Control & \shortstack{Pooled $\Delta$Elo {[95\% CI]}} & W/L & \shortstack{RMSE reduction {[95\% CI]}} \\
\midrule
\textbf{\method{} (full)} & \textit{Reference} & -- & \textit{Reference} \\
\midrule
Global baseline & 295.7 [209.6, 393.9] & 46/12 & 0.00819 [0.00235, 0.01388] \\
Two-global-TFM ensemble & 209.6 [111.1, 321.3] & 44/14 & 0.00547 [$-$0.00032, 0.01093] \\
$k$NN residual & 112.0 [37.9, 191.7] & 41/17 & 0.00402 [$-$0.00110, 0.00930] \\
Single-molecule residual & 160.9 [78.7, 253.9] & 43/15 & 0.00768 [0.00251, 0.01295] \\
Shuffled relation labels & 88.9 [14.0, 168.8] & 41/17 & 0.00405 [$-$0.00100, 0.00947] \\
No reliability gate & 325.2 [235.3, 439.8] & 45/13 & 0.03443 [0.01947, 0.04956] \\
\bottomrule
\end{tabular*}
\end{table}

Table~\ref{tab:design-studies-main} shows that \method{} wins 46 of 58 tasks against the global baseline, 44 against the two-global-TFM ensemble, 41 against $k$NN residual, and 43 against single-molecule residual.
All four comparisons favor \method{} in pooled Elo.
Against single-molecule residual, the mean MoleculeACE RMSE reduction is 0.00768, whereas the intervals against the ensemble and $k$NN residual include zero.
Molecular comparisons therefore improve task-level outcomes over the evaluated simpler alternatives, although the RMSE advantage is less conclusive in some comparisons.

\subsection{Do the pair labels and reliability gate improve predictions?}
\label{sec:component-results}
\label{app:independent-controls}

\method{} outperforms shuffled relation labels on 41 of 58 tasks, with a pooled Elo advantage of 88.9 (Table~\ref{tab:design-studies-main}).
Matching molecular pairs with their correct targets therefore improves task-level performance, although the MoleculeACE RMSE-reduction interval includes zero.
The reliability gate has the largest effect among the evaluated alternatives: \method{} outperforms the ungated variant on 45 of 58 tasks and achieves a 0.03443 lower mean RMSE, the largest RMSE difference among these comparisons.
Appendix~\ref{app:fixed-gate-results} compares the default similarity-adaptive reliability gate with fixed coefficients shared across queries.

\subsection{What is the tradeoff between predictive performance and runtime?}
\label{sec:performance-cost}

\begin{wrapfigure}[17]{R}{0.6\textwidth}
\vspace{-1.4em}
\centering
\includegraphics[
trim=25 5 0 0 ,
width=0.9\linewidth]
{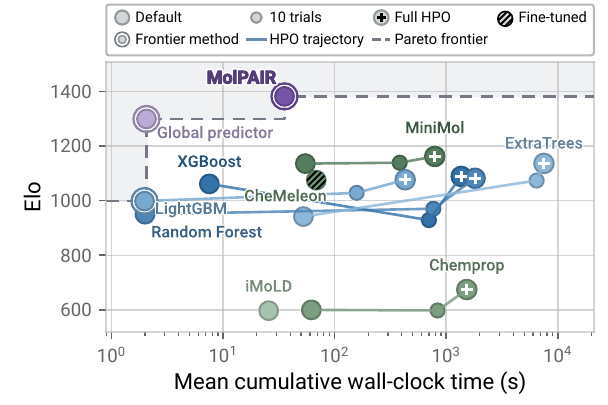}
\caption{\textbf{Elo versus computational cost.} Higher Elo and lower wall-clock time are better (upper left).}
\label{fig:performance-cost}
\end{wrapfigure}

The predictive gains of \method{} require additional computation for OOF preparation and molecular-pair inference, raising the question of whether these gains justify the added cost.
Figure~\ref{fig:performance-cost} compares predictive performance with mean cumulative wall-clock time across 58 structural-generalization tasks.
Elo is fitted jointly across 22 configurations including default baselines, while runtime includes hyperparameter search, fitting, and inference.
The connected points trace each method from its default to 10-trial and full HPO.

As expected, \method{} takes more time than its global predictor.
However, it achieves the highest Elo while requiring less time than every supervised method with even a 10-trial hyperparameter search, placing it on the observed Pareto frontier.
Thus, the additional computation for molecular-pair correction provides a favorable performance--cost tradeoff compared with task-specific hyperparameter tuning.
Appendix~\ref{app:cost-results} separately reports the time required to prepare each method and to predict properties for new molecules.

\section{Conclusion}
\label{sec:discussion}
\label{sec:conclusion}

We investigate tabular in-context learning for structural generalization in molecular property prediction and introduce \method{} to improve it through explicit molecular comparisons.
\method{} combines a global TFM that predicts properties from molecule-level examples with a second TFM that predicts error differences between a query and labeled reference molecules. %
The global baseline ranks second among the evaluated methods, establishing a competitive starting point.
\method{} improves this baseline on 46 of 58 tasks, with gains across all twelve representation--backbone combinations and alternative structural split protocols.
Additional analyses further show that the gains cannot be explained simply by using additional labeled molecules or a second predictor.
These results support explicit molecular-pair correction as a useful augmentation to tabular in-context learning for structural generalization in molecular property prediction.

\mypar{Limitations and future work.}
\method{} increases inference cost, and some component comparisons show clear task-level gains without conclusive improvements in native RMSE.
Broader real-world distribution shifts may also involve changes that are not captured by scaffold separation and fingerprint-similarity-based splits.
Future work can therefore evaluate the approach under broader molecular shifts, develop more efficient pair selection and inference, and improve the reliability and uncertainty estimates used to determine when molecular comparison should be applied.

\clearpage

\bibliography{references}

@article{tynes2021pairwise,
  title={Pairwise difference regression: a machine learning meta-algorithm for improved prediction and uncertainty quantification in chemical search},
  author={Tynes, Michael and Gao, Wenhao and Burrill, Daniel J and Batista, Enrique R and Perez, Danny and Yang, Ping and Lubbers, Nicholas},
  journal={Journal of chemical information and modeling},
  volume={61},
  number={8},
  pages={3846--3857},
  year={2021},
  publisher={ACS Publications}
}

@article{hollmann2025accurate,
  title={Accurate predictions on small data with a tabular foundation model},
  author={Hollmann, Noah and M{\"u}ller, Samuel and Purucker, Lennart and Krishnakumar, Arjun and K{\"o}rfer, Max and Hoo, Shi Bin and Schirrmeister, Robin Tibor and Hutter, Frank},
  journal={Nature},
  volume={637},
  number={8045},
  pages={319--326},
  year={2025},
  publisher={Nature Publishing Group UK London}
}

@article{grinsztajn2026tabpfn,
  title={Tabpfn-3: Technical report},
  author={Grinsztajn, L{\'e}o and Fl{\"o}ge, Klemens and Key, Oscar and Birkel, Felix and Jund, Philipp and Roof, Brendan and Manium, Mihir and Hoo, Shi Bin and B{\"u}hler, Magnus and Garg, Anurag and others},
  journal={arXiv preprint arXiv:2605.13986},
  year={2026}
}

@article{qu2026tabiclv2,
  title={TabICLv2: A better, faster, scalable, and open tabular foundation model},
  author={Qu, Jingang and Holzm{\"u}ller, David and Varoquaux, Ga{\"e}l and Le Morvan, Marine},
  journal={arXiv preprint arXiv:2602.11139},
  year={2026}
}

@article{qu2025tabicl,
  title={Tabicl: A tabular foundation model for in-context learning on large data},
  author={Qu, Jingang and Holzm{\"u}ller, David and Varoquaux, Ga{\"e}l and Le Morvan, Marine},
  journal={arXiv preprint arXiv:2502.05564},
  year={2025}
}

@article{hicham2026tabular,
  title={Tabular foundation models for in-context prediction of molecular properties},
  author={Ben Hicham, Karim K. and Rittig, Jan G and Grohe, Martin and Mitsos, Alexander},
  journal={arXiv preprint arXiv:2604.16123},
  year={2026}
}

@article{guan2026can,
  title={Can Tabular In-Context Learners Generalize to Biomolecular Property Prediction?},
  author={Guan, Davy and Zhang, Lu and Wijesinghe, Asiri and Zhu, Allen and Zhao, He and Power, Helen and Ahmed, F Hafna and Warden, Andrew and Ong, Cheng Soon and Steinberg, Daniel M},
  journal={arXiv preprint arXiv:2606.31126},
  year={2026}
}

@article{erickson2026tabarena,
  title={Tabarena: A living benchmark for machine learning on tabular data},
  author={Erickson, Nick and Purucker, Lennart and Tschalzev, Andrej and Holzm{\"u}ller, David and Desai, Prateek and Salinas, David and Hutter, Frank},
  journal={Advances in Neural Information Processing Systems},
  volume={38},
  year={2026}
}

@article{burns2025deep,
  title={Deep Learning Foundation Models from Classical Molecular Descriptors},
  author={Burns, Jackson W and Zalte, Akshat Shirish and Abreu, Charlles RA and Sieg, Jochen and Feldmann, Christian and Mathea, Miriam and Green, William H},
  journal={arXiv preprint arXiv:2506.15792},
  year={2025}
}

@article{klaser2404minimol,
  title={{MiniMol}: A parameter-efficient foundation model for molecular learning},
  author={Kl{\"a}ser, Kerstin and Banaszewski, B{\l}azej and Maddrell-Mander, Samuel and McLean, Callum and M{\"u}ller, Luis and Parviz, Ali and Huang, Shenyang and Fitzgibbon, Andrew},
  journal={arXiv preprint arXiv:2404.14986},
  year={2024}
}

@article{banaszewski2026monroe,
  title={Monroe: A Molecular Foundation Model for In-Context Probabilistic Inference},
  author={Banaszewski, Blazej and Fitzgibbon, Andrew W},
  journal={arXiv preprint arXiv:2608.18982},
  year={2026}
}

@article{yang2019analyzing,
  title={Analyzing learned molecular representations for property prediction},
  author={Yang, Kevin and Swanson, Kyle and Jin, Wengong and Coley, Connor and Eiden, Philipp and Gao, Hua and Guzman-Perez, Angel and Hopper, Timothy and Kelley, Brian and Mathea, Miriam and others},
  journal={Journal of chemical information and modeling},
  volume={59},
  number={8},
  pages={3370},
  year={2019}
}

@article{wang2022molecular,
  title={Molecular contrastive learning of representations via graph neural networks},
  author={Wang, Yuyang and Wang, Jianren and Cao, Zhonglin and Barati Farimani, Amir},
  journal={Nature Machine Intelligence},
  volume={4},
  number={3},
  pages={279--287},
  year={2022},
  publisher={Nature Publishing Group UK London}
}

@article{ross2022large,
  title={Large-scale chemical language representations capture molecular structure and properties},
  author={Ross, Jerret and Belgodere, Brian and Chenthamarakshan, Vijil and Padhi, Inkit and Mroueh, Youssef and Das, Payel},
  journal={Nature Machine Intelligence},
  volume={4},
  number={12},
  pages={1256--1264},
  year={2022},
  publisher={Nature Publishing Group UK London}
}

@article{fralish2023deepdelta,
  title={DeepDelta: predicting ADMET improvements of molecular derivatives with deep learning},
  author={Fralish, Zachary and Chen, Ashley and Skaluba, Paul and Reker, Daniel},
  journal={Journal of cheminformatics},
  volume={15},
  number={1},
  pages={101},
  year={2023},
  publisher={Springer}
}

@article{van2022exposing,
  title={Exposing the limitations of molecular machine learning with activity cliffs},
  author={Van Tilborg, Derek and Alenicheva, Alisa and Grisoni, Francesca},
  journal={Journal of chemical information and modeling},
  volume={62},
  number={23},
  pages={5938},
  year={2022}
}

@misc{polaris2026,
  title = {Polaris: The Benchmarking Platform for Drug Discovery},
  author = {{Polaris Hub}},
  year = {2026},
  howpublished = {\url{https://polarishub.io/}},
  note = {Accessed August 2026}
}

@article{rogers2010extended,
  title={Extended-connectivity fingerprints},
  author={Rogers, David and Hahn, Mathew},
  journal={Journal of chemical information and modeling},
  volume={50},
  number={5},
  pages={742--754},
  year={2010},
  publisher={ACS Publications}
}

@article{bemis1996properties,
  title={The properties of known drugs. 1. Molecular frameworks},
  author={Bemis, Guy W and Murcko, Mark A},
  journal={Journal of medicinal chemistry},
  volume={39},
  number={15},
  pages={2887--2893},
  year={1996},
  publisher={ACS Publications}
}

@article{hussain2010computationally,
  title={Computationally efficient algorithm to identify matched molecular pairs (MMPs) in large data sets},
  author={Hussain, Jameed and Rea, Ceara},
  journal={Journal of chemical information and modeling},
  volume={50},
  number={3},
  pages={339--348},
  year={2010},
  publisher={ACS Publications}
}

@article{joeres2025data,
  title={Data splitting to avoid information leakage with DataSAIL},
  author={Joeres, Roman and Blumenthal, David B and Kalinina, Olga V},
  journal={Nature communications},
  volume={16},
  number={1},
  pages={3337},
  year={2025},
  publisher={Nature Publishing Group UK London}
}

@article{steshin2023hi,
  title={Lo-hi: Practical ml drug discovery benchmark},
  author={Steshin, Simon},
  journal={Advances in Neural Information Processing Systems},
  volume={36},
  pages={64526--64554},
  year={2023}
}

@article{yang2022learning,
  title={Learning substructure invariance for out-of-distribution molecular representations},
  author={Yang, Nianzu and Zeng, Kaipeng and Wu, Qitian and Jia, Xiaosong and Yan, Junchi},
  journal={Advances in Neural Information Processing Systems},
  volume={35},
  pages={12964--12978},
  year={2022}
}

@article{zhuang2023learning,
  title={Learning invariant molecular representation in latent discrete space},
  author={Zhuang, Xiang and Zhang, Qiang and Ding, Keyan and Bian, Yatao and Wang, Xiao and Lv, Jingsong and Chen, Hongyang and Chen, Huajun},
  journal={Advances in Neural Information Processing Systems},
  volume={36},
  pages={78435--78452},
  year={2023}
}

@inproceedings{ji2023drugood,
  title={Drugood: Out-of-distribution dataset curator and benchmark for ai-aided drug discovery--a focus on affinity prediction problems with noise annotations},
  author={Ji, Yuanfeng and Zhang, Lu and Wu, Jiaxiang and Wu, Bingzhe and Li, Lanqing and Huang, Long-Kai and Xu, Tingyang and Rong, Yu and Ren, Jie and Xue, Ding and others},
  booktitle={Proceedings of the AAAI Conference on Artificial Intelligence},
  volume={37},
  pages={8023--8031},
  year={2023}
}

@misc{rdkit2024,
  title = {{RDKit}: Open-Source Cheminformatics},
  author = {Landrum, Greg and others},
  year = {2024},
  howpublished = {\url{https://www.rdkit.org/}}
}

@article{breiman2001random,
  title={Random forests},
  author={Breiman, Leo},
  journal={Machine learning},
  volume={45},
  number={1},
  pages={5--32},
  year={2001},
  publisher={Springer}
}

@article{geurts2006extremely,
  title={Extremely randomized trees},
  author={Geurts, Pierre and Ernst, Damien and Wehenkel, Louis},
  journal={Machine learning},
  volume={63},
  number={1},
  pages={3--42},
  year={2006},
  publisher={Springer}
}

@inproceedings{chen2016xgboost,
  title={Xgboost: A scalable tree boosting system},
  author={Chen, Tianqi and Guestrin, Carlos},
  booktitle={Proceedings of the 22nd acm sigkdd international conference on knowledge discovery and data mining},
  pages={785--794},
  year={2016}
}

@article{ke2017lightgbm,
  title={Lightgbm: A highly efficient gradient boosting decision tree},
  author={Ke, Guolin and Meng, Qi and Finley, Thomas and Wang, Taifeng and Chen, Wei and Ma, Weidong and Ye, Qiwei and Liu, Tie-Yan},
  journal={Advances in neural information processing systems},
  volume={30},
  year={2017}
}

@inproceedings{akiba2019optuna,
  title={Optuna: A next-generation hyperparameter optimization framework},
  author={Akiba, Takuya and Sano, Shotaro and Yanase, Toshihiko and Ohta, Takeru and Koyama, Masanori},
  booktitle={Proceedings of the 25th ACM SIGKDD international conference on knowledge discovery \& data mining},
  pages={2623--2631},
  year={2019}
}

@article{bradley1952rank,
  title={Rank analysis of incomplete block designs: I. the method of paired comparisons},
  author={Bradley, Ralph Allan and Terry, Milton E},
  journal={Biometrika},
  volume={39},
  number={3/4},
  pages={324--345},
  year={1952},
  publisher={JSTOR}
}

@article{wu2018moleculenet,
  title={MoleculeNet: a benchmark for molecular machine learning},
  author={Wu, Zhenqin and Ramsundar, Bharath and Feinberg, Evan N and Gomes, Joseph and Geniesse, Caleb and Pappu, Aneesh S and Leswing, Karl and Pande, Vijay},
  journal={Chemical science},
  volume={9},
  number={2},
  pages={513--530},
  year={2018},
  publisher={The Royal Society of Chemistry}
}

@article{wallach2018most,
  title={Most ligand-based classification benchmarks reward memorization rather than generalization},
  author={Wallach, Izhar and Heifets, Abraham},
  journal={Journal of chemical information and modeling},
  volume={58},
  number={5},
  pages={916--932},
  year={2018},
  publisher={ACS Publications}
}

@article{sheridan2013time,
  title={Time-split cross-validation as a method for estimating the goodness of prospective prediction.},
  author={Sheridan, Robert P},
  journal={Journal of chemical information and modeling},
  volume={53},
  number={4},
  pages={783--790},
  year={2013},
  publisher={ACS Publications}
}

@article{landrum2023simpd,
  title={SIMPD: an algorithm for generating simulated time splits for validating machine learning approaches},
  author={Landrum, Gregory A and Beckers, Maximilian and Lanini, Jessica and Schneider, Nadine and Stiefl, Nikolaus and Riniker, Sereina},
  journal={Journal of cheminformatics},
  volume={15},
  number={1},
  pages={119},
  year={2023},
  publisher={Springer}
}

@article{hollmann2022tabpfn,
  title={Tabpfn: A transformer that solves small tabular classification problems in a second},
  author={Hollmann, Noah and M{\"u}ller, Samuel and Eggensperger, Katharina and Hutter, Frank},
  journal={arXiv preprint arXiv:2207.01848},
  year={2022}
}

@article{schneider2006scaffold,
  title={Scaffold-hopping: how far can you jump?},
  author={Schneider, Gisbert and Schneider, Petra and Renner, Steffen},
  journal={QSAR \& Combinatorial Science},
  volume={25},
  number={12},
  pages={1162--1171},
  year={2006},
  publisher={Wiley Online Library}
}

@article{muller2021transformers,
  title={Transformers can do bayesian inference},
  author={M{\"u}ller, Samuel and Hollmann, Noah and Arango, Sebastian Pineda and Grabocka, Josif and Hutter, Frank},
  journal={arXiv preprint arXiv:2112.10510},
  year={2021}
}

@misc{cho2026causilotechnicalreport,
  title={{Causilo Technical Report}},
  author={Minyong Cho and Minho Jeong and Dooho Lee and Jinmo Lee and Jaemin Yoo},
  year={2026},
  eprint={2609.22866},
  archivePrefix={arXiv},
  primaryClass={cs.LG},
  url={https://arxiv.org/abs/2609.22866}
}
\bibliographystyle{iclr2027_conference}

\clearpage
\appendix
\section*{Appendix contents}
\begingroup
\small
\newcommand{\appentry}[2]{\noindent\hyperref[#1]{\ref*{#1}\quad #2}\dotfill\pageref*{#1}\par}
\newcommand{\appsubentry}[2]{\noindent\hspace*{1em}\hyperref[#1]{\ref*{#1}\quad #2}\dotfill\pageref*{#1}\par}
\appentry{app:method}{Implementation details}
\appsubentry{app:implementation-settings}{Preprocessing and inference settings}
\appsubentry{app:construction-properties}{Cross-fitting and correction targets}
\appentry{app:experiments}{Datasets and experimental settings}
\appsubentry{app:datasets}{Dataset versions and metrics}
\appsubentry{app:primary-splits}{Split construction and retained populations}
\appsubentry{app:alternative-split-protocols}{Independent partitions and split fingerprints}
\appsubentry{app:baseline-settings}{Representations, checkpoints, and tuning}
\appsubentry{app:specialized-rankings}{Specialized baseline configurations}
\appsubentry{app:shared-global-protocol}{Prediction-strategy controls}
\appsubentry{app:aggregation}{Aggregation and comparison pools}
\appsubentry{app:runtime-protocol}{Runtime accounting}
\appentry{app:results}{Supplementary prediction comparisons}
\appsubentry{app:specialized-results}{Comparison with specialized molecular methods}
\appsubentry{app:new-series-metrics}{Native metrics and seed variation}
\appsubentry{app:representation-backbone-results}{Encoder and backbone robustness}
\appsubentry{app:shared-global-results}{Single-molecule controls and pair targets}
\appsubentry{app:standard-mechanisms}{Cross-fitting, anchor choice, and aggregation on standard splits}
\appsubentry{app:relation-design-results}{Relation-design controls}
\appsubentry{app:fixed-gate-results}{Fixed versus similarity-adaptive gating}
\appentry{app:other-settings}{Evaluation beyond the primary structural splits}
\appsubentry{app:scaffold-controls}{Prediction strategies on scaffold-only splits}
\appsubentry{app:lohi-results}{Hit identification and lead optimization}
\appsubentry{app:external-shifts}{Assay, size, and chronological shifts}
\appentry{app:diagnostics}{Prediction changes and serving costs}
\appsubentry{app:cached-support}{Query-level improvements and fallback}
\appsubentry{app:activity-cliff-results}{Predictions on activity-cliff molecules}
\appsubentry{app:cost-results}{Setup, label updates, and repeated query costs}
\appentry{app:reproducibility}{Reproducibility and result provenance}
\appsubentry{app:software}{Software and hardware}
\appsubentry{app:integrity}{Data integrity and execution}
\endgroup
\clearpage
\section{Implementation Details}
\label{app:method}

The method first predicts a molecular property from a molecule-level context, then uses a second context of molecular pairs to adjust that prediction.
This section specifies how those two inputs and their prediction targets are constructed, so that the changes tested in the supplementary experiments can be understood relative to one reference configuration.
The primary configuration uses CheMeleon with TabPFN-3 and RDKit2D relation features.
All relation-context pairs are selected within the same OOF fold, including in the supplementary evaluations.

\subsection{Preprocessing and inference settings}
\label{app:implementation-settings}

\begin{table}[htbp]
\centering
\small
\caption{\textbf{Default context and inference settings.}}
\label{tab:appendix-settings}
\begin{tabular}{@{}ll@{}}
\toprule
Setting & Value \\
\midrule
Global representation & CheMeleon, 2,048 input features before filtering \\
Relation representation & RDKit2D, up to 996 selected features \\
Global and relation backbone & TabPFN-3, eight estimators \\
Cross-fitting & Five folds within the outer training split \\
Training anchors & Up to eight non-self anchors within the same OOF fold \\
Relation context budget & At most 8,192 rows, deterministic subsampling \\
Query anchors & At most 32 training molecules \\
Anchor fingerprint & Morgan radius two, Tanimoto similarity \\
Anchor priority & Exact one-cut matched-pair cores first \\
Query anchor acceptance & Similarity at least 0.20 \\
Reliability threshold & $\tau=0.20$ \\
Relation prediction batches & At most 8,192 directed pair rows \\
Relation estimator seed & Global estimator seed plus 104,729 \\
\bottomrule
\end{tabular}
\end{table}

The outer training molecules are the labeled examples available for the prediction task; queries are the held-out molecules whose properties are to be predicted.
All preprocessing is fitted on the outer training molecules and then applied to queries, preventing the held-out data from determining input transformations.
We use median imputation, constant-feature filtering, and variance selection.
The global table retains at most 2,000 features.
The relation table contains signed and absolute molecular differences plus eight auxiliary signals, leaving $(2000-8)/2=996$ molecular features.
RDKit2D relations use filtered descriptors without \ac{PCA}.

Table~\ref{tab:appendix-settings} summarizes the default context and inference settings.
The global predictor uses a regressor or classifier according to the endpoint.
The relation predictor always uses a regressor because error differences are continuous, including for binary endpoints.
At query time, anchor predictions and uncertainties remain their stored cross-fitted values, while the query uses the full training context.
Both pair directions are evaluated before aggregating anchor corrections.
Context construction creates preprocessing state and a reusable in-context estimator, but does not update the pretrained checkpoints.

\subsection{Cross-fitting and correction targets}
\label{app:construction-properties}

\paragraph{Cross-fitting.}
The relation model needs examples of errors made on labeled molecules, but an error computed after giving the global model that molecule's own label can be misleadingly small.
We therefore divide training molecules into folds and predict each fold using the remaining folds as context; these are out-of-fold (OOF) predictions.
Excluding molecule $i$'s fold from its prediction context prevents its label $y_i$ from directly conditioning the prediction used to define $e_i$.
With OOF predictions, both molecules in a pair belong to the same held-out fold, so neither label conditions the global predictions used for that pair.
The in-sample-residual ablation replaces these held-out predictions but retains the pair-selection rule.
Residuals can still be dependent because the prediction contexts overlap across folds.

\paragraph{Pair direction and chemical support.}
The difference between two errors changes sign when the two molecules exchange roles.
The antisymmetric combination in the main method guarantees $\widetilde d(b,a)=-\widetilde d(a,b)$ even if the relation predictor does not learn this property.
A self-pair therefore has zero predicted difference.
An anchor is a labeled training molecule selected as a reference for a query.
The gate reduces the adjustment when even the most similar accepted anchor has low structural similarity to the query.
Its value approaches zero as that maximum similarity approaches $\tau$.
With no accepted anchor, the method returns the global prediction.

\paragraph{Correction targets.}
The same pair of molecules can supply different learning targets: the difference between their measured properties, or the difference between the errors in their global predictions.
Distinguishing these targets is necessary to interpret the supplementary target comparison.
Direct property-difference prediction estimates $y_b-y_a$ without global-prediction features.
Property-delta correction retains those features but uses the same property-difference target and forms a proposal from the anchor label.
Residual-difference correction instead predicts $(y_b-\mu_b^{-})-(y_a-\mu_a^{-})$ and refines the query's global prediction.
Only the latter two preserve global-prediction information, so a comparison with direct property-difference prediction does not isolate the target choice.
Appendix~\ref{app:correction-target-results} evaluates the two correction targets with shared global outputs.

\paragraph{Uncertainty for weighting and final diagnostics.}
Uncertainty is used for two separate purposes: weighting individual anchors and summarizing uncertainty in the final prediction.
These quantities are not interchangeable.
The relation predictor's scale $u^R_{aq}$ enters the anchor weights through similarity squared divided by this scale, with the numerical floors in Equation~\ref{eq:anchor-weights}.
This pair-level scale is distinct from the dispersion of the correction candidates.
For accepted anchors, let $\bar c_q=\sum_a w_{aq}c_{aq}$.
The correction-uncertainty summary is the weighted root-mean-square dispersion
\begin{equation}
    v_q=\left[\sum_a w_{aq}(c_{aq}-\bar c_q)^2\right]^{1/2}.
\end{equation}
For regression, the final uncertainty scale used in the diagnostics is the global scale plus this dispersion, $u_q+v_q$.
In the pairwise branch, the reliability gate multiplies the point correction $\bar c_q$, but does not multiply $v_q$.
The resulting scale is a heuristic summary, not an exact posterior uncertainty or a guarantee of calibrated coverage.
In particular, shrinking the point correction need not shrink the reported uncertainty.

\section{Datasets and Experimental Settings}
\label{app:experiments}

This section explains how the experiments translate structural generalization into train--test comparisons, how the competing methods are fitted, and which quantities are held fixed in the additional controls.
We distinguish a change to the evaluation population from a change to the prediction method, because these answer different questions.
Alternative structural partitioning is specified in Appendix~\ref{app:alternative-split-protocols}; official and chronological benchmark settings are given with their detailed results in Appendix~\ref{app:other-settings}.

\subsection{Dataset versions and metrics}
\label{app:datasets}

Each task asks for one property or endpoint to be predicted from molecular structure.
Combining the two suites lets us examine multiple regression and classification problems rather than one property alone.
Table~\ref{tab:datasets} summarizes the benchmark suites, planned seeds, and released primary metrics.
Using each task's released metric keeps the comparisons aligned with the task's original evaluation objective.

\begin{table}[!htbp]
\centering
\small
\caption{\textbf{Benchmark tasks and metrics.} Planned seeds are per endpoint.}
\label{tab:datasets}
\begin{tabular}{@{}lrrp{0.18\textwidth}p{0.35\textwidth}@{}}
\toprule
Suite & Endpoints & Planned seeds & Task types & Primary metrics \\
\midrule
MoleculeACE & 30 & 5 & regression & \acs*{RMSE} \\
Polaris & 28 & 5 & mixed & \acs*{MAE}, \acs*{MSE}, Pearson, Spearman, \acs*{PR-AUC}, \acs*{ROC-AUC} \\
\midrule
Total & 58 & 5 & mixed & released metric \\
\bottomrule
\end{tabular}
\end{table}

MoleculeACE is pinned to repository revision \texttt{7e6de0bd2968c56589c5\allowbreak 80f2a397f01c531ede26}.
Polaris uses \texttt{polaris-lib==0.13.0}.
The pipeline checks every benchmark and underlying dataset against its declared checksum.
It assigns stable row identifiers before feature generation and checks cached row order to keep features aligned with labels.

MoleculeACE contains 30 ChEMBL affinity or activity endpoints and evaluates \ac{RMSE} on the provided splits~\citep{van2022exposing}.
Polaris contributes 28 curated endpoints with their released metrics.
Four use \ac{MAE}, three use \ac{MSE}, six use Pearson correlation, four use \ac{PR-AUC}, seven use \ac{ROC-AUC}, and four use Spearman correlation.
Released target transforms and evaluators are retained for every method in each comparison.

\subsection{Split construction and retained populations}
\label{app:primary-splits}

\paragraph{What the split is intended to test.}
The primary experiment asks whether a model predicts well for molecules whose structures are not closely represented in its labeled training set.
Holding out arbitrary individual molecules does not guarantee this separation: a test molecule can still have a close structural analogue in training.
We therefore first hold out entire Bemis--Murcko scaffold groups and then remove test molecules that remain too similar to any training molecule.
The scaffold grouping and fingerprint filter play different roles: the former separates scaffold groups, while the latter checks remaining structural similarity across the partition.
This operational definition is not a claim that every retained molecule belongs to a chemically unrelated family.

We distinguish three protocols throughout the paper.
\emph{Standard splits} are the partitions released by the benchmarks.
\emph{Scaffold-only splits} hold out complete scaffold groups without the additional similarity filter.
The \emph{primary structural-generalization protocol} combines scaffold-group separation with the maximum-training-similarity cutoff.
Results from these protocols are kept separate because they need not evaluate the same molecules or the same degree of structural separation.

\paragraph{Construction and applicability.}
The structural-shift experiments construct new partitions from the released training and validation rows, without using the released benchmark test labels.
For scaffold-only evaluation, the splitter considers 128 deterministic group-split candidates with zero scaffold overlap and selects the one closest to a 20\% test fraction.
The primary protocol instead starts from a 25\% scaffold-group holdout.
For each candidate test molecule, it computes Morgan Tanimoto similarity to the training molecules and retains the molecule only if its maximum similarity does not exceed the selected cutoff.
A task--seed split is recorded as not applicable when fewer than 16 test molecules remain.

At the default cutoff of 0.60, 286 of 290 planned splits are applicable and yield 55,733 query--seed predictions.
The task \texttt{CHEMBL2835\_Ki} retains one seed; all other tasks retain five.
Methods in each paired comparison use exactly the same retained row identifiers.
The audit records why rows or splits were excluded and the maximum training similarity of each retained test molecule.

\paragraph{Why vary the cutoff?}
A gain observed only at 0.60 could depend on that particular test population.
We therefore repeat the comparison at cutoffs from 0.40 to 0.70, allowing fewer or more structurally similar test molecules to remain.
Table~\ref{tab:split-audit} reports the resulting populations and direct Elo gains; Table~\ref{tab:similarity-threshold-robustness} reports task wins and MoleculeACE RMSE.
\method{} wins a majority of tasks at each cutoff, but the mean RMSE reduction at 0.40 is small.
Because stricter cutoffs remove molecules and sometimes entire task--seed splits, differences between rows are not a controlled trajectory of the same population as novelty increases.

The upper evaluation cutoff must not be confused with the lower anchor-acceptance threshold.
The former determines which molecules are tested; the latter determines which labeled molecules \method{} may use as references for a given query.
In this experiment, both the anchor-acceptance threshold and reliability-gate threshold remain fixed at 0.20.
Thus, we change the evaluation population, not the rule used to produce a prediction.

\begin{table}[t]
\centering
\footnotesize
\setlength{\tabcolsep}{3.2pt}
\caption{\textbf{Retained populations by similarity cutoff.} Valid splits are out of 290 planned task--seed combinations. Prediction counts include repeats across seeds. Positive direct $\Delta$Elo favors \method{} over the global predictor.}
\label{tab:split-audit}
\begin{tabular}{rrrrr}
\toprule
Max. similarity & Valid splits & Query--seed predictions & Endpoints & Direct $\Delta$Elo \\
\midrule
0.40 & 219 / 290 & 26,488 & 52 & 171.9 \\
0.50 & 268 / 290 & 42,159 & 54 & 197.7 \\
0.60 & 286 / 290 & 55,733 & 58 & 231.1 \\
0.70 & 289 / 290 & 72,474 & 58 & 166.3 \\
\bottomrule
\end{tabular}
\end{table}

\subsection{Independent partitions and split fingerprints}
\label{app:alternative-split-protocols}
\label{app:new-series-extensions}
\label{app:cutoff-results}
\label{app:fingerprint-results}

\paragraph{Purpose.}
Changing the cutoff still retains our scaffold-based partitioning procedure.
To test whether the gain depends on that procedure, we also use DataSAIL C1 to construct 80:20 molecular partitions from pairwise similarity~\citep{joeres2025data}.
We use either ECFP4 or MACCS fingerprints to define similarity for partitioning.
Separately, we compare Morgan- and MACCS-based anchor selection within \method{}.
The split fingerprint determines which molecules enter training and test; the anchor fingerprint determines which training molecules are chosen as references after the split is fixed.
Keeping these choices distinct lets us ask whether improvement requires using the same structural representation for both operations.

\paragraph{Compared configurations.}
For each partitioning scheme, we evaluate the global predictor, \method{} with RDKit2D relation features and Morgan anchors, and the same relation model with MACCS anchors.
The primary CheMeleon--TabPFN-3 configuration is retained.
Each three-method Elo fit is performed separately within its partitioning scheme; ratings from different schemes should not be read as a ranking of split difficulty.
Across the two split fingerprints, the study covers 58 tasks, five seeds, and three configurations, yielding 1,740 completed runs.
The ECFP4 scheme has 290 applicable task--seed splits, 69,025 test predictions, and a median test-set size of 154.5.

\paragraph{Results and validation.}
With Morgan anchor selection unchanged, \method{} wins 38 of 58 tasks against the global predictor under each partitioning scheme.
Its pooled Elo contrast is +73.8 [12.2, 138.0] on ECFP4 partitions and +111.5 [33.5, 190.6] on MACCS partitions.
The benefit therefore appears beyond the primary scaffold-plus-cutoff construction and does not require matching the split and anchor fingerprints.
This does not show that every alternative anchor rule is equally effective; the reported contrasts here concern the primary Morgan-anchor configuration.

\subsection{Representations, checkpoints, and tuning}
\label{app:baseline-settings}

\paragraph{Representations and checkpoints.}
The global predictor and the relation predictor receive different kinds of input: molecule-level representations for the former and pairwise features for the latter.
We record the encoder and TFM checkpoints so that changes in the model input or pretrained weights are not confused with changes to the context design.
Table~\ref{tab:representations} lists the molecular representations and their input dimensions before train-only filtering.

\begin{table}[!htbp]
\centering
\small
\caption{\textbf{Molecular representations before train-only filtering.}}
\label{tab:representations}
\begin{tabular}{lrl}
\toprule
Representation & Dimension & Source or use \\
\midrule
RDKit2D & data dependent & deterministic descriptors \\
Morgan & 1,024 to 4,096 & radius two or three, bit or count \\
CheMeleon & 2,048 & frozen directed message-passing fingerprint \\
MiniMol & 512 & frozen pretrained \acs*{GINE} fingerprint \\
Monroe & released dimension & frozen pretrained molecular fingerprint \\
\bottomrule
\end{tabular}
\end{table}

TabPFN uses the released version-three classifier and regressor checkpoints.
TabICLv2 uses the classifier and regressor checkpoints released on 2026-02-12.
Both models use eight estimators, and the pipeline verifies each checkpoint file by SHA-256 before planning a job.
Causilo provides the third frozen backbone in the main robustness experiment~\citep{cho2026causilotechnicalreport}; checkpoint identity and inference settings are recorded with each run.

\paragraph{Tree-model selection.}
To compare with task-trained tree predictors, we allow both their molecular representation and model hyperparameters to be selected on training data.
The inner validation folds guide selection; the outer test set measures performance only after that selection is complete.
Every search uses only the current outer training partition; the held-out molecules are never used to select a representation, hyperparameter, or boosting iteration.
The objective is the endpoint's released primary metric in its declared direction.
The search jointly selects one of 25 representations and the estimator parameters.
The representations comprise RDKit2D and every Morgan radius in $\{2,3\}$, fingerprint type in $\{\text{bit},\text{count}\}$, and dimension in $\{1024,2048,4096\}$, both alone and concatenated with RDKit2D.
Preprocessing and feature filtering are refitted within every shuffled inner fold, using stratification for classification when possible.

A multi-fidelity screen evaluates all 25 representations with three folds, retains ten for a medium-budget three-fold screen, and retains five for a full-budget five-fold screen.
The three stages use 256, 512, and 1,000 trees for forest models and at most 1,000, 2,000, and 5,000 rounds for boosting models.
Optuna then searches the best three representations with three folds~\citep{akiba2019optuna}.
The three best trials and the best locked-default configuration receive a final five-fold audit at full budget.
Random Forest and ExtraTrees use 30 Optuna trials; XGBoost and LightGBM use 50.
Table~\ref{tab:new-series-tree-hpo} gives the model-specific hyperparameter search spaces.

\begin{table*}[t]
\centering
\scriptsize
\setlength{\tabcolsep}{4pt}
\caption{\textbf{Tree-baseline search spaces.} Log denotes logarithmic sampling.}
\label{tab:new-series-tree-hpo}
\begin{tabular}{@{}lp{0.78\textwidth}@{}}
\toprule
Model and trials & Search space \\
\midrule
Random Forest, 30 & \texttt{max\_features} in $\{\sqrt{d},\log_2 d,0.2,0.5,0.8,1.0\}$; \texttt{max\_depth} in $\{\texttt{None},8,16,32,64\}$; \texttt{min\_samples\_leaf} from 1 to 16 (log); \texttt{min\_samples\_split} from 2 to 32 (log); bootstrap on or off; and, for classification, class weight in \{none, balanced, balanced subsample\}. \\
ExtraTrees, 30 & The same space as Random Forest. \\
XGBoost, 50 & Learning rate from 0.005 to 0.2 (log), depth 3--12, child weight 0.01--20 (log), row subsampling 0.5--1.0, column subsampling 0.4--1.0, gamma 0--5, L1 $10^{-8}$--10 (log), L2 $10^{-3}$--100 (log), and maximum bins in $\{128,256,512\}$. \\
LightGBM, 50 & Learning rate 0.005--0.2 (log), leaves 15--255 (log), depth in $\{-1,4,6,8,12,16\}$, minimum child rows 5--100 (log), row subsampling 0.5--1.0, column subsampling 0.4--1.0, L1 $10^{-8}$--10 (log), L2 $10^{-3}$--100 (log), and maximum bins in $\{127,255,511\}$. \\
\bottomrule
\end{tabular}
\end{table*}

Boosting uses early-stopping patience 50 during search and 100 during the final audit.
The selected iteration count is the median best iteration across the five audit folds, after which the selected representation and estimator are fitted once on the complete outer training partition.
The search seed is 20260827.
The cache key contains the exact outer-training row identifiers but excludes the evaluation seed, so identical outer partitions reuse one selection while distinct structural-generalization partitions are searched separately.
The final fitted estimator uses the evaluation seed.
The main benchmark retains the tuned variants of these methods. The runtime analysis additionally includes their default and 10-trial variants.

\paragraph{Chemprop, MiniMol, and CheMeleon.}
These methods represent different ways to fit a molecular predictor: training a model end to end, fitting a head on frozen embeddings, or fine-tuning a pretrained encoder.
Their budgets are specified separately because they do not share the tree models' representation search.
Chemprop and MiniMol each use 20 multivariate TPE trials and one train-only 80:20 holdout, stratified for classification when class counts permit.
The sampler and holdout seed are 20260910 plus the evaluation seed, and the cache key contains the evaluation seed; these searches therefore run independently for every outer evaluation seed.
Table~\ref{tab:new-series-molecular-hpo} gives the search spaces for Chemprop and the MiniMol prediction head.

\begin{table*}[t]
\centering
\scriptsize
\setlength{\tabcolsep}{4pt}
\caption{\textbf{Molecular-baseline search spaces.} Log denotes logarithmic sampling.}
\label{tab:new-series-molecular-hpo}
\begin{tabular}{@{}lp{0.78\textwidth}@{}}
\toprule
Model and trials & Search space \\
\midrule
Chemprop, 20 & Message width in $\{256,512,768,1024\}$, message-passing depth 3--6, mean or sum aggregation, feed-forward width in $\{128,256,512,1024\}$, one to three feed-forward layers, dropout 0--0.4 in steps of 0.1, batch size in $\{32,64,128\}$, one to five warmup epochs, maximum learning rate $3\times10^{-5}$--$3\times10^{-3}$ (log), initial-rate ratio 0.02--0.2 (log), final-rate ratio 0.001--0.05 (log), and class balancing on or off for classification. \\
MiniMol head, 20 & Hidden-width multiplier in $\{0.5,1,2\}$, one to three hidden layers, dropout 0--0.3 in steps of 0.1, learning rate $3\times10^{-5}$--$2\times10^{-3}$ (log), weight decay $10^{-7}$--$10^{-3}$ (log), batch size in $\{16,32,64,128\}$, epochs in $\{15,25,40\}$, and warmup epochs in $\{2,5,8\}$. \\
\bottomrule
\end{tabular}
\end{table*}

Chemprop trials run for at most 20 epochs with patience three.
The selected configuration trains a new model for at most 50 epochs with patience five on a separately seeded train-only 80:20 split, rather than refitting one network on all outer-training rows.
MiniMol keeps its 512-dimensional encoder frozen; each trial uses a three-fold head ensemble, and the selected setting is trained as a five-fold ensemble over the complete outer training partition.
The canonical CheMeleon recipe uses a train-only 80:20 split, at most 50 epochs, and patience five.
The main comparison retains this canonical CheMeleon configuration.
Monroe is used as a frozen representation in the encoder--backbone study, without task-specific encoder tuning.

\subsection{Specialized baseline configurations}
\label{app:specialized-protocol}
\label{app:specialized-rankings}

These baselines ask whether frozen tabular inference is competitive with methods designed for molecular pairs or molecular distribution shifts.
PADRE and DeepDelta supply the pairwise comparison; iMoLD and MoleOOD supply task-trained distribution-shift methods.
The specialized comparisons use the same outer structural-generalization partitions and evaluation seeds as the primary methods.
None uses an outer test label for training, early stopping, or model selection.
PADRE and DeepDelta are evaluated on 47 regression endpoints, MoleOOD on 11 classification endpoints, and iMoLD on both.
The applicable run counts are 231, 55, and 286 for regression-only, classification-only, and full-scope methods, respectively.

\paragraph{PADRE with Random Forest.}
Because the published PADRE source is not publicly executable, this baseline implements the public pseudocode associated with DOI \texttt{10.1021/acs.jcim.1c00670}.
It uses 2,048-bit radius-two Morgan fingerprints and ordered molecule pairs, including self-pairs, with target $y_b-y_a$ and input $[x_a,x_b]$.
We retain all pairs up to a deterministic cap of 65,536 per endpoint and seed.
The fixed random forest has 50 trees, minimum leaf size one, and \texttt{max\_features}=1.0, with no HPO.
At inference, each query uses at most 512 Morgan-similar training anchors, and its absolute prediction is the mean of $y_a+\widehat{\Delta}_{a,q}$ across anchors.

\paragraph{DeepDelta.}
We pin the official repository at revision \texttt{cd9b131} and implement its shared two-molecule D-MPNN objective through Chemprop 2.3~\citep{fralish2023deepdelta}.
No pretrained checkpoint is used.
The model is trained from scratch for every regression endpoint and seed using sum aggregation, a train-only 80:20 molecule split, five epochs, patience five, and batch size 128, without HPO.
Training and validation are capped at 65,536 and 8,192 pairs, respectively.
For absolute Core58 predictions, each query uses at most 128 Morgan-similar training anchors and averages $y_a+\widehat{\Delta}_{a,q}$.

\paragraph{iMoLD.}
We pin the official implementation at revision \texttt{6d24452} and port its objective and graph feature contract to Core58.
Released checkpoints for other datasets are not used.
Every endpoint and seed trains from scratch with a 20\% train-only validation split, 4,000 codebook entries, embedding dimension 128, four layers, dropout 0.5, $\gamma=0.9$, invariance weight 0.01, regularization weight 0.5, learning rate 0.001, and batch size 128.
The fixed budget is 200 epochs with patience 30 and no HPO.
Checkpoint selection uses validation RMSE for regression and validation ROC-AUC for classification.

\paragraph{MoleOOD.}
We pin the official implementation at revision \texttt{5bf49ab} and port its objective to the 11 binary endpoints under the primary protocol~\citep{yang2022learning}.
Released checkpoints for other endpoints are not used.
Every run trains from scratch with BRICS decomposition, 20 latent domains, a 20\% stratified train-only validation split, 20 assistant epochs, 50 main epochs, learning rate 0.001, batch size 128, and loss weight one.
The configuration is fixed without HPO, and the main checkpoint is selected by validation ROC-AUC.

Appendix~\ref{app:specialized-results} reports these comparisons in Table~\ref{tab:specialized-method-comparison}.

\subsection{Prediction-strategy controls}
\label{app:shared-global-protocol}

\paragraph{Prediction errors.}
In this subsection, a residual means the observed label minus the corresponding global prediction; an OOF residual uses a global prediction made without that molecule's fold in context.

\paragraph{The seven configurations in the main comparison.}
Table~\ref{tab:design-studies-main} contains \method{}, the global baseline, the two-global-TFM ensemble, $k$NN residual transfer, single-molecule residual prediction, shuffled relation labels, and the no-gate variant.
They use the primary structural-generalization partitions.
The $k$NN, single-molecule, shuffled-label, and no-gate controls reuse the same global predictions as \method{}.
Elo is fitted jointly over all seven, and each displayed contrast subtracts the comparator's rating from the \method{} rating within that fit.

The $k$NN alternative uses stored errors from similar labeled molecules without learning pairwise error differences.
The ensemble instead combines two global predictions, with no RDKit2D relation context, anchors, residual targets, or reliability gate.
Its TabPFN-3 models use seeds $s$ and $s+104729$, the same checkpoint, and eight estimators each.
Both receive the same frozen CheMeleon features and train-only preprocessing as the primary global predictor.
Regression predictions and corresponding quantiles are averaged; classification probabilities are averaged before thresholding at 0.5 when a class label is needed.
Its diagnostic uncertainty is $(u_1+u_2)/2+|\hat y_1-\hat y_2|/2$.

\paragraph{Single-molecule controls.}
Appendix~\ref{app:shared-global-results} reports a regression/classification breakdown and a context-size control for single-molecule prediction.
The gated single-residual model predicts individual OOF errors and uses the same reliability gate as \method{}.
The row-matched model also matches the relation table's row count and input width, but not the cost of bidirectional inference.
Both retain CheMeleon--TabPFN-3.
The pair model uses same-fold OOF pairs, while the single-molecule models predict the stored individual OOF errors directly.
The row-matched comparison tests whether table dimensions alone explain the gain.

\paragraph{Pair targets and scaffold-only evaluation.}
The property-delta alternative keeps the molecular pairs and global-output input features, but predicts the measured property difference rather than the difference between global prediction errors.
It forms a query proposal from the anchor's label.
The target comparison and the scaffold-only controls reuse saved global features, predictions, five-fold OOF outputs, uncertainties, and molecule order.
They retain the primary encoder and backbone and use same-fold molecular pairs.
Their four-method pool consists of \method{}, the gated single-residual alternative, the row-matched single-residual alternative, and property-delta prediction; the standalone global baseline is not included.
Results for the primary protocol and scaffold-only partitions are reported separately in Appendices~\ref{app:shared-global-results} and~\ref{app:scaffold-controls}.

\paragraph{Additional design variants.}
Appendix~\ref{app:relation-design-results} changes input features, anchor weights, anchor counts, gate thresholds, and matched-pair priority while retaining the reference's saved global and OOF artifacts.
It includes ten variants across the same 58 tasks and 286 applicable splits.
The full reference reuses its corresponding completed results.
A separate contrast in this set compares pair prediction with $k$NN transfer when both use similarity-only weighting and the same global outputs.
This comparison uses similarity-only weights for both methods, rather than the full method's similarity-and-uncertainty weights.
All eleven reported contrasts are direct two-method fits, not differences from the seven-method Elo pool.

\paragraph{Checks on released standard splits.}
The cross-fitting, random-anchor, bidirectional-symmetry, and uncertainty-weighting controls in Appendix~\ref{app:standard-mechanisms} use the benchmarks' released standard splits.
They execute global predictors independently and are not part of the primary structural-generalization comparison.
All molecular-pair variants retain same-fold pair selection.
In particular, the in-sample variant changes which global predictions are used to compute errors, not the pair-selection rule.
Table~\ref{tab:standard-component-controls} reports task wins and native-metric differences for these experiments.

\subsection{Aggregation and comparison pools}
\label{app:aggregation}

Different tasks have different metric scales, so a large numeric improvement on one task should not automatically outweigh many smaller improvements on others.
We therefore report both original task metrics and task-level comparisons, with each task receiving equal weight in the benchmark summaries.
Here an endpoint is one prediction task, not one molecule.

\paragraph{Native metrics and endpoint wins.}
For each endpoint, we first average each method's native metric across the available seeds.
These seed-averaged scores determine the reported endpoint W/L/T, after accounting for whether the metric is minimized or maximized.
Cross-endpoint native-metric summaries also give each endpoint equal weight.

\paragraph{A common endpoint-first Elo rule.}
All Elo analyses use the seed-averaged endpoint scores defined above.
For each method pair and endpoint, the better score receives one win, the worse score one loss, and a tie contributes one half to each method, accounting for the direction of the native metric.
Each endpoint has unit weight per method pair, regardless of its number of available seeds or test molecules.
This rule is identical for direct two-method fits, the ten-method leaderboard, the 22-configuration Pareto analysis, and all shared-control and protocol-specific pools.
Individual seed wins are not treated as separate Elo observations.
With the same fitting settings, identical endpoint W/L/T counts yield identical direct two-method contrasts: \method{} has W/L of 44/3 against both PADRE and DeepDelta, giving the same \method{}-minus-comparator contrast of 451.9 [298.1, 809.9].

Each Elo fit uses the outcomes and weights specified above in a penalized Bradley and Terry model.
For methods $i$ and $j$,
\begin{equation}
P(i>j)=\frac{1}{1+\exp[-(\beta_i-\beta_j)]}.
\end{equation}
The fitted coefficients are converted to Elo ratings as
\begin{equation}
R_i=1000+\frac{400}{\ln 10}\beta_i,
\end{equation}
after centering them so that the mean rating is 1000 within the evaluated method pool.

We obtain 95\% confidence intervals from 10,000 endpoint-bootstrap samples.
Each sample draws complete endpoints with replacement and refits the Bradley and Terry model from their endpoint-level pairwise outcomes.
Rating intervals use the 2.5th and 97.5th percentiles, while contrast intervals use paired rating differences within each bootstrap sample.

Although the aggregation rule is shared, Elo ratings and contrasts depend on the methods included in the fit.
The main benchmark table uses a ten-method pool, while the performance--cost figure uses 22 configurations.
Direct comparisons fit only the two indicated methods.
The prediction-and-correction contrasts in Table~\ref{tab:design-studies-main} use all seven configurations displayed in that table.
Table~\ref{tab:standard-component-controls} reports W/L counts and native-metric reductions for the standard-split component study, rather than Elo contrasts from its separate comparison pool.
The shared-output structural-generalization and scaffold controls each use the four configurations named in Appendix~\ref{app:shared-global-protocol}.
DataSAIL fingerprint contrasts use three-method pools fitted separately for ECFP4 and MACCS partitions.
Displaying only selected contrasts does not refit those pools.
A direct two-method contrast refits just that pair, whereas a pooled contrast also depends on the other comparisons in its pool.
The same pairwise W/L/T can therefore yield different pooled contrasts, and contrasts from different pools are not directly comparable.
Compatible native metrics are averaged with equal endpoint weight, and their paired differences use the same endpoint-bootstrap principle.

\subsection{Runtime accounting}
\label{app:runtime-protocol}

The main runtime question concerns the total computation used to obtain predictions for a task, including model selection when applicable, rather than only the speed of an already prepared predictor.
We compare 58 structural-generalization tasks using a fixed pool of 22 configurations.
These comprise default, 10-trial, and full-HPO versions of Random Forest, ExtraTrees, XGBoost, LightGBM, Chemprop, and the MiniMol head, together with iMoLD, canonical CheMeleon fine-tuning, the primary global predictor, and \method{}.
Default costs include fitting and inference.
Ten-trial costs include the first ten search trials followed by fitting and inference with the selected configuration; full-HPO costs include the complete search and final fitting and inference.
The \method{} workflow additionally includes OOF preparation, relation-context construction, and query--anchor inference.

The horizontal axis reports mean cumulative wall-clock time on a logarithmic scale.
Within-method lines follow the observed HPO budget sequence, which need not improve performance monotonically.
A configuration is Pareto-dominated when another has at least as high Elo and no greater runtime, with a strict improvement in at least one dimension.
The separate setup and query-batch measurements in Appendix~\ref{app:cost-results} diagnose serving costs and are not pooled with this end-to-end comparison.

\section{Supplementary Prediction Comparisons}
\label{app:results}

The main experiments compare complete prediction methods. Here we give their task-level results and examine more closely why adding molecular comparisons can help.
Unless a subsection states otherwise, the reference is the primary \method{} configuration: a CheMeleon--TabPFN-3 global predictor, an RDKit2D relation context, and molecular pairs formed within the same held-out OOF fold.
An anchor is a labeled training molecule used as a reference for predicting a query; an OOF error is computed from a prediction made without using that molecule's fold as context.
In each experiment below, we specify which parts of this reference are retained and which are changed.

\subsection{Comparison with specialized molecular methods}
\label{app:specialized-results}
\label{sec:specialized-results}

\paragraph{Purpose and comparison.}
The main benchmark includes general-purpose predictors and molecular models, but does not by itself show how \method{} compares with methods specifically designed to learn from molecular pairs or handle molecular distribution shifts.
We therefore evaluate PADRE (RF) and DeepDelta on the 47 regression tasks, MoleOOD on the 11 classification tasks, and iMoLD on both~\citep{tynes2021pairwise,fralish2023deepdelta,yang2022learning,zhuang2023learning}.
Unlike \method{}'s frozen relation TFM, PADRE and DeepDelta fit task-specific models to molecular property differences.
This comparison asks whether our use of a frozen pair predictor is competitive with these evaluated task-trained alternatives; it does not isolate the effect of pairwise inputs alone.

All methods use the same applicable primary structural-generalization partitions and evaluation seeds.
Their training uses only the outer training data, with the fixed budgets and implementations specified in Appendix~\ref{app:specialized-protocol}.
We report task wins across each method's supported task types, and mean RMSE on the 30 MoleculeACE tasks.
Direct Elo contrasts fit \method{} and one comparator at a time.

\begin{table}[htbp]
\centering
\small
\setlength{\tabcolsep}{5pt}
\caption{\textbf{Comparison with specialized molecular methods.} W/L and direct $\Delta$Elo compare \method{} with each method. RMSE uses MoleculeACE; positive reductions and Elo favor \method{}. Brackets give 95\% intervals; -- denotes unavailable or inapplicable values.}
\label{tab:specialized-method-comparison}
\scalebox{0.92}{%
\begin{tabular}{@{}lcccc@{}}
\toprule
Method & MACE RMSE & \shortstack{RMSE reduction {[95\% CI]}} & W/L & \shortstack{Direct $\Delta$Elo {[95\% CI]}} \\
\midrule
\multicolumn{5}{l}{\textit{Regression on 47 tasks}} \\
\method{} & 0.8849 & Reference & -- & -- \\
Global baseline & 0.8931 & 0.00828 [0.00246, 0.01427] & 40/7 & 298.1 [184.0, 451.9] \\
PADRE (RF) & 1.0012 & 0.11632 [0.08687, 0.14614] & 44/3 & 451.9 [298.1, 809.9] \\
DeepDelta & 1.0588 & 0.17393 [0.14058, 0.20757] & 44/3 & 451.9 [298.1, 809.9] \\
iMoLD & 1.0796 & 0.19478 [0.15370, 0.24499] & 47/0 & 809.9 \\
\midrule
\multicolumn{5}{l}{\textit{Classification on 11 tasks}} \\
Global baseline & -- & -- & 6/5 & -- \\
iMoLD & -- & -- & 11/0 & -- \\
MoleOOD & -- & -- & 11/0 & -- \\
\bottomrule
\end{tabular}
}
\end{table}

\paragraph{Results.}
Table~\ref{tab:specialized-method-comparison} shows that \method{} wins 44 of 47 regression tasks against both PADRE (RF) and DeepDelta, and all 47 against iMoLD.
It also wins all 11 classification tasks against both MoleOOD and iMoLD.
On MoleculeACE, its mean RMSE is 0.8849, compared with 1.0012 for PADRE (RF) and 1.0588 for DeepDelta.
Thus, learning task-specific pair-model parameters is not necessary to obtain competitive predictions in this comparison.
The conclusion concerns these implementations and budgets, not every possible tuning of the specialized methods.
\FloatBarrier

\subsection{Native metrics and seed variation}
\label{app:new-series-metrics}
\label{app:endpoint-details}

Elo and task win counts summarize which method performs better, but do not show how large the difference is on a particular task.
Tables~\ref{tab:new-series-mace-endpoints} and~\ref{tab:new-series-polaris-endpoints} therefore report the same primary-protocol results in each task's original evaluation metric.
For each task and method, the tables summarize performance across its available evaluation seeds; they are a more detailed view of the benchmark, not an additional experiment.
The released target transformations and metric directions are unchanged.
MoleculeACE tasks share RMSE, whereas the Polaris results retain their different error and correlation measures rather than combining incompatible scales.
The single applicable seed for \texttt{CHEMBL2835\_Ki} does not provide a standard deviation.

\begin{table}[!htbp]
\centering
\scriptsize
\setlength{\tabcolsep}{2.2pt}
\caption{\textbf{MoleculeACE endpoint RMSE.} Values are mean $\pm$ SD across available seeds. No SD is available for the single-seed endpoint \texttt{CHEMBL2835\_Ki}.}
\label{tab:new-series-mace-endpoints}
\resizebox{\textwidth}{!}{%
\begin{tabular}{@{}llrrrrr@{}}
\toprule
Endpoint & Metric & \method{} & Global & MiniMol & XGBoost & Chemprop \\
\midrule
1862 Ki & \acs*{RMSE} $\downarrow$ & $\mathbf{0.741\pm0.104}$ & $0.747\pm0.102$ & $0.844\pm0.116$ & $0.751\pm0.058$ & $1.028\pm0.125$ \\
1871 Ki & \acs*{RMSE} $\downarrow$ & $\mathbf{0.770\pm0.102}$ & $0.774\pm0.099$ & $0.802\pm0.105$ & $0.815\pm0.093$ & $0.806\pm0.075$ \\
2034 Ki & \acs*{RMSE} $\downarrow$ & $\mathbf{0.940\pm0.066}$ & $0.963\pm0.068$ & $0.996\pm0.038$ & $1.067\pm0.040$ & $1.188\pm0.102$ \\
2047 EC50 & \acs*{RMSE} $\downarrow$ & $0.934\pm0.135$ & $0.915\pm0.109$ & $\mathbf{0.881\pm0.130}$ & $0.960\pm0.118$ & $0.912\pm0.066$ \\
204 Ki & \acs*{RMSE} $\downarrow$ & $\mathbf{0.904\pm0.174}$ & $0.911\pm0.169$ & $1.022\pm0.103$ & $0.981\pm0.092$ & $1.077\pm0.100$ \\
2147 Ki & \acs*{RMSE} $\downarrow$ & $\mathbf{0.695\pm0.066}$ & $0.704\pm0.077$ & $0.791\pm0.052$ & $0.827\pm0.121$ & $0.944\pm0.111$ \\
214 Ki & \acs*{RMSE} $\downarrow$ & $\mathbf{0.879\pm0.067}$ & $0.883\pm0.061$ & $0.930\pm0.056$ & $0.905\pm0.089$ & $0.936\pm0.083$ \\
218 EC50 & \acs*{RMSE} $\downarrow$ & $0.808\pm0.091$ & $0.816\pm0.105$ & $0.833\pm0.125$ & $\mathbf{0.804\pm0.135}$ & $0.983\pm0.092$ \\
219 Ki & \acs*{RMSE} $\downarrow$ & $0.940\pm0.084$ & $0.936\pm0.072$ & $\mathbf{0.908\pm0.052}$ & $0.946\pm0.050$ & $1.053\pm0.088$ \\
228 Ki & \acs*{RMSE} $\downarrow$ & $\mathbf{1.011\pm0.071}$ & $1.021\pm0.058$ & $1.090\pm0.062$ & $1.097\pm0.084$ & $1.141\pm0.075$ \\
231 Ki & \acs*{RMSE} $\downarrow$ & $\mathbf{0.788\pm0.138}$ & $0.797\pm0.139$ & $0.790\pm0.112$ & $0.866\pm0.180$ & $0.955\pm0.137$ \\
233 Ki & \acs*{RMSE} $\downarrow$ & $\mathbf{0.988\pm0.031}$ & $1.004\pm0.028$ & $0.995\pm0.051$ & $1.021\pm0.014$ & $1.065\pm0.054$ \\
234 Ki & \acs*{RMSE} $\downarrow$ & $\mathbf{0.912\pm0.095}$ & $0.913\pm0.085$ & $0.920\pm0.057$ & $0.946\pm0.054$ & $1.071\pm0.084$ \\
235 EC50 & \acs*{RMSE} $\downarrow$ & $\mathbf{0.877\pm0.069}$ & $0.884\pm0.070$ & $0.907\pm0.052$ & $0.882\pm0.091$ & $0.971\pm0.142$ \\
236 Ki & \acs*{RMSE} $\downarrow$ & $\mathbf{0.837\pm0.124}$ & $0.851\pm0.132$ & $0.909\pm0.140$ & $0.852\pm0.144$ & $1.036\pm0.187$ \\
237 EC50 & \acs*{RMSE} $\downarrow$ & $0.984\pm0.071$ & $\mathbf{0.973\pm0.055}$ & $1.075\pm0.050$ & $1.057\pm0.076$ & $1.151\pm0.267$ \\
237 Ki & \acs*{RMSE} $\downarrow$ & $1.057\pm0.126$ & $1.077\pm0.144$ & $1.105\pm0.161$ & $\mathbf{1.008\pm0.129}$ & $1.148\pm0.105$ \\
238 Ki & \acs*{RMSE} $\downarrow$ & $0.933\pm0.055$ & $\mathbf{0.903\pm0.052}$ & $0.920\pm0.066$ & $0.906\pm0.029$ & $1.053\pm0.140$ \\
239 EC50 & \acs*{RMSE} $\downarrow$ & $\mathbf{0.980\pm0.116}$ & $1.007\pm0.100$ & $1.002\pm0.171$ & $1.009\pm0.148$ & $1.183\pm0.122$ \\
244 Ki & \acs*{RMSE} $\downarrow$ & $1.016\pm0.092$ & $\mathbf{1.013\pm0.077}$ & $1.179\pm0.084$ & $1.035\pm0.083$ & $1.178\pm0.096$ \\
262 Ki & \acs*{RMSE} $\downarrow$ & $\mathbf{0.726\pm0.043}$ & $0.731\pm0.047$ & $0.731\pm0.022$ & $0.757\pm0.062$ & $0.885\pm0.091$ \\
264 Ki & \acs*{RMSE} $\downarrow$ & $\mathbf{0.908\pm0.069}$ & $0.920\pm0.073$ & $0.918\pm0.051$ & $0.967\pm0.027$ & $1.048\pm0.093$ \\
2835 Ki & \acs*{RMSE} $\downarrow$ & $\mathbf{0.726}$ (n/a) & $0.777$ (n/a) & $0.798$ (n/a) & $0.764$ (n/a) & $1.246$ (n/a) \\
287 Ki & \acs*{RMSE} $\downarrow$ & $\mathbf{0.899\pm0.048}$ & $0.915\pm0.044$ & $0.946\pm0.065$ & $0.917\pm0.047$ & $1.007\pm0.031$ \\
2971 Ki & \acs*{RMSE} $\downarrow$ & $\mathbf{0.979\pm0.134}$ & $0.984\pm0.117$ & $0.994\pm0.102$ & $1.055\pm0.149$ & $1.244\pm0.115$ \\
3979 EC50 & \acs*{RMSE} $\downarrow$ & $\mathbf{0.852\pm0.104}$ & $0.855\pm0.096$ & $0.877\pm0.079$ & $0.883\pm0.125$ & $0.943\pm0.161$ \\
4005 Ki & \acs*{RMSE} $\downarrow$ & $\mathbf{0.792\pm0.122}$ & $0.815\pm0.096$ & $0.938\pm0.104$ & $0.909\pm0.132$ & $1.055\pm0.111$ \\
4203 Ki & \acs*{RMSE} $\downarrow$ & $\mathbf{0.784\pm0.053}$ & $0.796\pm0.061$ & $0.826\pm0.042$ & $0.814\pm0.053$ & $0.947\pm0.098$ \\
4616 EC50 & \acs*{RMSE} $\downarrow$ & $0.871\pm0.089$ & $0.908\pm0.093$ & $0.929\pm0.128$ & $\mathbf{0.868\pm0.134}$ & $1.065\pm0.213$ \\
4792 Ki & \acs*{RMSE} $\downarrow$ & $1.022\pm0.158$ & $\mathbf{1.000\pm0.131}$ & $1.072\pm0.150$ & $1.057\pm0.192$ & $1.075\pm0.254$ \\
\bottomrule
\end{tabular}}
\end{table}

\begin{table}[!htbp]
\centering
\fontsize{7.5}{9}\selectfont
\setlength{\tabcolsep}{1.8pt}
\caption{\textbf{Polaris endpoint metrics.} Values are mean $\pm$ SD across available seeds; arrows indicate the preferred direction.}
\label{tab:new-series-polaris-endpoints}
\begin{tabular*}{\textwidth}{@{\extracolsep{\fill}}>{\raggedright\arraybackslash}p{0.19\textwidth}>{\raggedright\arraybackslash}p{0.10\textwidth}*{5}{>{\centering\arraybackslash}p{0.13\textwidth}}@{}}
\toprule
Endpoint & Metric & \method{} & Global & MiniMol & XGBoost & Chemprop \\
\midrule
adme fang hclint 1 & Pearson $\uparrow$
& $0.696\pm0.010$ & $0.673\pm0.012$ & $\mathbf{0.734\pm0.029}$ & $0.689\pm0.023$ & $0.630\pm0.039$ \\

adme fang hppb 1 & Pearson $\uparrow$
& $0.605\pm0.109$ & $0.587\pm0.106$ & $\mathbf{0.723\pm0.118}$ & $0.614\pm0.122$ & $0.226\pm0.228$ \\

adme fang perm 1 & Pearson $\uparrow$
& $0.700\pm0.035$ & $0.674\pm0.038$ & $\mathbf{0.762\pm0.040}$ & $0.713\pm0.038$ & $0.641\pm0.033$ \\

adme fang rclint 1 & Pearson $\uparrow$
& $0.717\pm0.008$ & $0.703\pm0.008$ & $\mathbf{0.737\pm0.012}$ & $0.711\pm0.019$ & $0.676\pm0.031$ \\

adme fang rppb 1 & Pearson $\uparrow$
& $0.594\pm0.072$ & $0.550\pm0.096$ & $\mathbf{0.665\pm0.108}$ & $0.513\pm0.080$ & $0.246\pm0.369$ \\

adme fang solu 1 & Pearson $\uparrow$
& $0.582\pm0.052$ & $0.565\pm0.047$ & $\mathbf{0.679\pm0.024}$ & $0.519\pm0.060$ & $0.538\pm0.072$ \\

pkis2 egfr wt reg v2 & \acs*{MSE} $\downarrow$
& \shortstack{$\mathbf{453.182}$\\$\mathbf{\pm74.826}$}
& \shortstack{$463.193$\\$\pm88.454$}
& \shortstack{$485.340$\\$\pm136.673$}
& \shortstack{$594.315$\\$\pm229.713$}
& \shortstack{$782.250$\\$\pm165.020$} \\

pkis2 kit wt cls v2 & \acs*{PR-AUC} $\uparrow$
& $0.656\pm0.091$ & $\mathbf{0.668\pm0.077}$ & $0.642\pm0.165$ & $0.569\pm0.097$ & $0.388\pm0.151$ \\

pkis2 kit wt reg v2 & \acs*{MSE} $\downarrow$
& \shortstack{$\mathbf{920.150}$\\$\mathbf{\pm67.110}$}
& \shortstack{$950.165$\\$\pm70.715$}
& \shortstack{$1040.120$\\$\pm380.662$}
& \shortstack{$964.783$\\$\pm103.319$}
& \shortstack{$1891.203$\\$\pm984.857$} \\

pkis2 ret wt cls v2 & \acs*{PR-AUC} $\uparrow$
& $0.559\pm0.087$ & $\mathbf{0.584\pm0.112}$ & $0.509\pm0.150$ & $0.488\pm0.179$ & $0.175\pm0.113$ \\

pkis2 ret wt reg v2 & \acs*{MSE} $\downarrow$
& \shortstack{$703.269$\\$\pm267.779$}
& \shortstack{$\mathbf{700.870}$\\$\mathbf{\pm218.448}$}
& \shortstack{$782.490$\\$\pm205.801$}
& \shortstack{$843.996$\\$\pm179.931$}
& \shortstack{$918.496$\\$\pm195.618$} \\

ames & \acs*{ROC-AUC} $\uparrow$
& $\mathbf{0.834\pm0.022}$ & $0.831\pm0.022$ & $0.826\pm0.011$ & $0.827\pm0.011$ & $0.816\pm0.019$ \\

bbb martins & \acs*{ROC-AUC} $\uparrow$
& $\mathbf{0.915\pm0.024}$ & $0.910\pm0.026$ & $0.886\pm0.025$ & $0.896\pm0.021$ & $0.824\pm0.057$ \\

bioavailability ma & \acs*{ROC-AUC} $\uparrow$
& $\mathbf{0.729\pm0.072}$ & $0.725\pm0.067$ & $0.726\pm0.043$ & $0.716\pm0.073$ & $0.674\pm0.069$ \\

caco2 wang & \acs*{MAE} $\downarrow$
& $\mathbf{0.399\pm0.010}$ & $0.405\pm0.009$ & $0.439\pm0.044$ & $0.404\pm0.023$ & $0.501\pm0.129$ \\

clearance hepatocyte az & Spearman $\uparrow$
& $0.413\pm0.072$ & $0.399\pm0.057$ & $\mathbf{0.421\pm0.086}$ & $0.329\pm0.092$ & $0.306\pm0.109$ \\

clearance microsome az & Spearman $\uparrow$
& $\mathbf{0.568\pm0.066}$ & $0.558\pm0.051$ & $0.561\pm0.026$ & $0.468\pm0.070$ & $0.387\pm0.108$ \\

cyp2c9 substrate carbonmangels & \acs*{PR-AUC} $\uparrow$
& $0.257\pm0.036$ & $0.276\pm0.023$ & $\mathbf{0.306\pm0.042}$ & $0.234\pm0.054$ & $0.195\pm0.027$ \\

cyp2d6 substrate carbonmangels & \acs*{PR-AUC} $\uparrow$
& $\mathbf{0.613\pm0.063}$ & $0.585\pm0.053$ & $0.537\pm0.066$ & $0.562\pm0.084$ & $0.455\pm0.076$ \\

cyp3a4 substrate carbonmangels & \acs*{ROC-AUC} $\uparrow$
& $0.692\pm0.035$ & $\mathbf{0.694\pm0.034}$ & $0.656\pm0.029$ & $0.655\pm0.049$ & $0.614\pm0.026$ \\

dili & \acs*{ROC-AUC} $\uparrow$
& $\mathbf{0.876\pm0.027}$ & $0.874\pm0.022$ & $0.873\pm0.029$ & $0.837\pm0.048$ & $0.860\pm0.033$ \\

half life obach & Spearman $\uparrow$
& $\mathbf{0.419\pm0.102}$ & $0.415\pm0.112$ & $0.261\pm0.180$ & $0.252\pm0.092$ & $0.108\pm0.048$ \\

herg & \acs*{ROC-AUC} $\uparrow$
& $0.876\pm0.029$ & $0.877\pm0.023$ & $\mathbf{0.892\pm0.028}$ & $0.858\pm0.029$ & $0.815\pm0.085$ \\

ld50 zhu & \acs*{MAE} $\downarrow$
& $\mathbf{0.462\pm0.009}$ & $0.464\pm0.008$ & $0.496\pm0.006$ & $0.468\pm0.009$ & $0.513\pm0.017$ \\

lipophilicity astrazeneca & \acs*{MAE} $\downarrow$
& $0.604\pm0.024$ & $0.624\pm0.029$ & $\mathbf{0.521\pm0.027}$ & $0.585\pm0.024$ & $0.555\pm0.059$ \\

pgp broccatelli & \acs*{ROC-AUC} $\uparrow$
& $0.887\pm0.014$ & $0.885\pm0.015$ & $\mathbf{0.894\pm0.013}$ & $0.883\pm0.010$ & $0.775\pm0.062$ \\

ppbr az & \acs*{MAE} $\downarrow$
& $10.421\pm0.861$ & $10.588\pm0.865$ & $\mathbf{9.725\pm0.776}$ & $10.739\pm0.853$ & $11.174\pm0.879$ \\

vdss lombardo & Spearman $\uparrow$
& $\mathbf{0.655\pm0.049}$ & $0.651\pm0.045$ & $0.513\pm0.058$ & $0.582\pm0.067$ & $0.448\pm0.071$ \\
\bottomrule
\end{tabular*}
\end{table}

\FloatBarrier

\subsection{Encoder and backbone robustness}
\label{app:representation-backbone-results}

\paragraph{Purpose and setup.}
An improvement with CheMeleon and TabPFN-3 could reflect that particular combination rather than a generally useful way to supply molecular comparisons.
We therefore cross four global representations---RDKit2D, CheMeleon, MiniMol, and Monroe---with TabPFN-3, TabICLv2, and Causilo, as in Section~\ref{sec:representation-backbone-results}.
For each combination, we compare its global predictor with \method{} using the same representation and backbone.
The relation context continues to use RDKit2D features and same-fold OOF pairing, and both TFMs in \method{} use the indicated backbone.
Thus, the global representation and backbone vary across rows, while the source of the relation features remains fixed.

Table~\ref{tab:representation-robustness} adds native regression errors to the Elo comparisons in Figure~\ref{fig:new-series-headline}.
RMSE summaries first average seeds within each of the 30 MoleculeACE tasks, then report the mean and standard deviation across tasks.
Elo is fitted separately to the two methods within each encoder--backbone combination, using seed-averaged outcomes on all 58 tasks.
These are twelve separate two-method comparisons, not one joint ranking of all configurations.

\begin{table}[htbp]
\centering
\footnotesize
\setlength{\tabcolsep}{3pt}
\caption{\textbf{Representation and backbone robustness.} RMSE is task mean $\pm$ SD on MoleculeACE. Direct $\Delta$Elo uses all 58 tasks; positive values favor \method{}. Bold marks lower RMSE within each pair.}
\label{tab:representation-robustness}
\resizebox{0.90\textwidth}{!}{%
\begin{tabular*}{\textwidth}{@{\extracolsep{\fill}}llccr@{}}
\toprule
Backbone & Representation & Global RMSE & \method{} RMSE & $\Delta$Elo \\
\midrule
TabPFN-3 & RDKit2D
& $0.9298 \pm 0.1211$ & $\mathbf{0.9227 \pm 0.1228}$ & +197.2 \\
TabPFN-3 & CheMeleon
& $0.8932 \pm 0.0966$ & $\mathbf{0.8849 \pm 0.1004}$ & +231.1 \\
TabPFN-3 & MiniMol
& $0.9106 \pm 0.1120$ & $\mathbf{0.9047 \pm 0.1114}$ & +290.8 \\
TabPFN-3 & Monroe
& $0.8918 \pm 0.0917$ & $\mathbf{0.8861 \pm 0.0943}$ & +166.3 \\
\midrule
TabICLv2 & RDKit2D
& $0.9310 \pm 0.1163$ & $\mathbf{0.9226 \pm 0.1151}$ & +166.3 \\
TabICLv2 & CheMeleon
& $0.9076 \pm 0.1070$ & $\mathbf{0.8979 \pm 0.1054}$ & +269.4 \\
TabICLv2 & MiniMol
& $0.9115 \pm 0.1082$ & $\mathbf{0.8985 \pm 0.1110}$ & +314.1 \\
TabICLv2 & Monroe
& $0.8994 \pm 0.1022$ & $\mathbf{0.8904 \pm 0.1022}$ & +181.4 \\
\midrule
Causilo & RDKit2D
& $0.9100 \pm 0.1208$ & $\mathbf{0.9066 \pm 0.1219}$ & +110.7 \\
Causilo & CheMeleon
& $0.8753 \pm 0.1063$ & $\mathbf{0.8711 \pm 0.1101}$ & +166.3 \\
Causilo & MiniMol
& $0.8925 \pm 0.1061$ & $\mathbf{0.8856 \pm 0.1102}$ & +124.0 \\
Causilo & Monroe
& $0.8780 \pm 0.0963$ & $\mathbf{0.8749 \pm 0.0983}$ & +97.7 \\
\bottomrule
\end{tabular*}%
}
\end{table}

\paragraph{Results.}
Mean MoleculeACE RMSE decreases in all twelve combinations.
MiniMol--TabICLv2 has the largest mean reduction, from 0.9115 to 0.8985, and the largest direct Elo gain (+314.1).
For the primary CheMeleon--TabPFN-3 configuration, mean RMSE decreases from 0.8932 to 0.8849 and the direct Elo gain is +231.1.
The results show gains across the tested encoders and backbones, including the non-pretrained RDKit2D representation, rather than only in the primary configuration.
\FloatBarrier

\subsection{Single-molecule controls and pair targets}
\label{app:correction-target-results}
\label{sec:correction-target-results}
\label{app:shared-global-results}
\label{app:single-molecule-results}

\paragraph{Purpose.}
We ask separately whether learning from pairs is more useful than predicting individual errors, and whether the pair target should be an error difference or a property difference.

\paragraph{Single-molecule alternatives.}
The gated single-residual model predicts each molecule's OOF error directly and applies the same reliability gate as \method{}.
The row-matched single-residual model additionally matches the number of context rows and input features supplied to the relation TFM.
The latter comparison tests whether a larger input table, rather than information about molecular pairs, could explain the gain.
Matching these dimensions does not match the cost of \method{}'s bidirectional pair inference.
Both alternatives retain CheMeleon--TabPFN-3 and reuse the reference global predictions; the pair model retains RDKit2D features and same-fold OOF pairs.

\begin{table}[htbp]
\centering
\footnotesize
\setlength{\tabcolsep}{3pt}
\caption{\textbf{Molecular-pair versus single-molecule correction.} W/L counts \method{} wins/losses. RMSE reductions use MoleculeACE; direct $\Delta$Elo uses all 58 tasks. Brackets give 95\% intervals.}
\label{tab:single-molecule-controls}
\begin{tabular*}{\columnwidth}{@{\extracolsep{\fill}}lcccc@{}}
\toprule
Control
& \shortstack{Regression\\W/L}
& \shortstack{Classification\\W/L}
& \shortstack{RMSE reduction\\{[95\% CI]}}
& \shortstack{Direct $\Delta$Elo\\{[95\% CI]}} \\
\midrule
Gated single residual
& 37/10 & 6/5
& 0.00768 [0.00260, 0.01292]
& 181.4 [85.0, 290.8] \\
Row-matched single residual
& 41/6 & 6/5
& 0.00479 [0.00110, 0.00826]
& 249.7 [151.7, 368.8] \\
\bottomrule
\end{tabular*}
\vspace{-1.2em}
\end{table}

Table~\ref{tab:single-molecule-controls} shows that \method{} wins 37 of 47 regression tasks against the gated alternative and 41 against the row-matched alternative.
The corresponding mean MoleculeACE RMSE reductions are 0.00768 and 0.00479, with both confidence intervals above zero.
On classification tasks, it wins six of 11 against each alternative, so the advantage is less consistent.
These comparisons support learning from molecular pairs beyond directly predicting individual errors or merely increasing the input table's dimensions, with the clearest evidence on regression.

\paragraph{Error differences versus property differences.}
Even if molecular pairs are useful, the results above do not establish which quantity the pair model should predict.
The primary method learns the difference between the two molecules' global prediction errors.
The property-delta alternative instead learns their measured property difference and uses the anchor's label to form a query prediction.
We keep the selected molecular pairs and global-output input features, and reuse the saved global predictions, OOF outputs, uncertainties, and molecule order.
This retains global-prediction information in both methods while changing the pair target (Appendix~\ref{app:shared-global-protocol}).

Against property-delta prediction, \method{} wins 31 of 58 tasks: 28 of 47 regression tasks and three of 11 classification tasks.
Mean MoleculeACE RMSE is 0.88496 for \method{} and 0.88383 for property-delta prediction.
The paired comparator-minus-\method{} difference is $-0.00114$ [$-0.00593$, 0.00209], whose interval includes zero.
Thus, the advantage over the evaluated single-molecule alternatives does not establish error differences as a better pairwise target than property differences.
\FloatBarrier

\subsection{Cross-fitting, anchor choice, and aggregation on standard splits}
\label{app:standard-mechanisms}

\paragraph{Purpose and evaluation setting.}
\method{} needs errors to learn from, reference molecules to compare with, and a rule for combining pair predictions.
We examine whether the way these quantities are constructed matters.
These experiments use the benchmarks' released standard splits, not the scaffold-plus-similarity splits used for the primary structural-generalization results.
They retain CheMeleon--TabPFN-3, RDKit2D relation features, and same-fold pair selection.
Global predictors are executed independently, so the results compare complete configurations under this standard-split setting.

\paragraph{What changes in each alternative.}
The in-sample alternative computes errors from global predictions whose context includes the molecule's own label, rather than using the held-out OOF predictions.
It tests the value of learning from errors on molecules excluded from the prediction context; it does not change which molecules can form a pair.
Random anchors replace chemically selected reference molecules with randomly selected ones.
Removing bidirectional symmetry omits the rule that combines predictions for both orders of a molecular pair.
Removing uncertainty weighting omits the relation model's uncertainty factor when combining anchor contributions.
The latter alternatives test reference selection and aggregation, rather than the choice of molecular encoder.

\begin{table}[htbp]
\centering
\footnotesize
\setlength{\tabcolsep}{4pt}
\caption{\textbf{Correction controls on released standard splits.} W/L counts \method{} wins/losses across 58 tasks. Positive MoleculeACE RMSE reductions favor \method{}; brackets give 95\% intervals.}
\label{tab:standard-component-controls}
\begin{tabular*}{\textwidth}{@{\extracolsep{\fill}}lcr@{}}
\toprule
Control & W/L & RMSE reduction [95\% CI] \\
\midrule
In-sample residuals & 44/14 & 0.01547 [0.00839, 0.02274] \\
Random anchors & 41/17 & 0.00373 [-0.00015, 0.00763] \\
No bidirectional symmetry & 33/25 & 0.00042 [-0.00015, 0.00105] \\
No uncertainty weighting & 28/30 & 0.00003 [-0.00038, 0.00045] \\
\bottomrule
\end{tabular*}
\end{table}

\paragraph{Results and scope.}
Table~\ref{tab:standard-component-controls} shows that \method{} wins 44 of 58 tasks against in-sample errors and reduces mean MoleculeACE RMSE by 0.01547 [0.00839, 0.02274].
It also wins 41 tasks against random anchors, although that RMSE-reduction interval includes zero.
The intervals for removing bidirectional symmetry or uncertainty weighting include zero as well.
The strongest native-metric evidence here therefore concerns the use of held-out errors, while the individual benefits of the other choices are less conclusive.
Because these experiments use different train--test partitions, they should not be combined with the primary-protocol controls to rank component importance under structural generalization.
\FloatBarrier

\subsection{Relation-design controls}
\label{app:relation-design-results}

\paragraph{Purpose and fixed reference.}
The main configuration uses predictions from two TFMs, a limited set of anchors, and similarity- and uncertainty-based weighting.
We test whether each of these design choices is needed, and whether simply transferring nearby errors can match learned pair predictions under the same anchor-weighting rule.
All ten additional variants use CheMeleon--TabPFN-3, RDKit2D relation features, and same-fold OOF pairing on the same 58 primary tasks and 286 applicable task--seed splits.
The global and OOF predictions remain fixed across these variants.

The reference uses at most 32 query anchors, gate threshold $\tau=0.20$, similarity-and-uncertainty weighting, and priority for matched molecular pairs.
We retain this prespecified reference throughout; we do not replace it with whichever variant performs best on the test set.

\paragraph{Input features and weighting.}
The relation TFM receives both molecular information and outputs of the global model.
To test whether those outputs help, one variant removes the global predictions, their difference, and their uncertainty summaries from the relation input, retaining the molecular descriptors and chemical-relation signals.
This removes them only from the pair model's input: the final prediction still starts from the same global output.
Similarity-only weighting uses normalized squared structural similarity and omits the inverse relation-uncertainty factor.
Uniform weighting gives every accepted anchor equal weight.
These alternatives ask whether the extra information and weighting rules improve on simpler uses of the same selected references.

\paragraph{Number and selection of reference molecules.}
The anchor-count variants change the maximum number of query anchors from 32 to 8, 16, or 64.
They test whether additional references help or instead add less useful comparisons.
The gate variants change $\tau$ to 0.10 or 0.30, testing weaker or stronger attenuation of the predicted adjustment, while leaving the minimum anchor-acceptance similarity at 0.20.
Thus, changing the gate threshold does not change the separate minimum similarity required to accept an anchor.
The reference additionally prioritizes matched molecular pairs, which are related through a localized structural change.
Removing matched-pair priority removes the ranking preference for exact one-cut matched molecular pairs, testing whether this chemical preference helps beyond the other anchor-selection criteria.

\begin{table}[htbp]
\centering
\footnotesize
\setlength{\tabcolsep}{4pt}
\caption{\textbf{Relation-design sensitivity.}
RMSE averages 30 MoleculeACE tasks; W/T/L compares the full reference with each variant on 58 tasks.
Positive direct $\Delta$Elo favors the reference; brackets give 95\% endpoint-bootstrap intervals.}
\label{tab:relation-design-full}
\begin{tabular*}{\textwidth}{@{\extracolsep{\fill}}lrrr@{}}
\toprule
Variant & MACE RMSE & Full W/T/L & Full $-$ variant $\Delta$Elo [95\% CI] \\
\midrule
Full reference & 0.88496 & -- & -- \\
No global-output features & 0.88577 & 35/0/23 & +72.4 [$-$11.9, 166.3] \\
Pair predictor, similarity-only weights & 0.88505 & 24/0/34 & $-$60.1 [$-$151.7, 23.8] \\
$k$NN residual, similarity-only weights & 0.88899 & 41/0/17 & +151.7 [60.1, 269.4] \\
Uniform weights & 0.88639 & 41/0/17 & +151.7 [60.1, 249.7] \\
\midrule
8 query anchors & 0.88436 & 27/0/31 & $-$23.8 [$-$110.7, 60.1] \\
16 query anchors & 0.88422 & 20/0/38 & $-$110.7 [$-$213.7, $-$23.8] \\
64 query anchors & 0.88572 & 38/2/18 & +124.0 [35.8, 231.1] \\
\midrule
Gate threshold 0.10 & 0.88542 & 24/0/34 & $-$60.1 [$-$151.7, 23.8] \\
Gate threshold 0.30 & 0.88609 & 41/0/17 & +151.7 [60.1, 269.4] \\
No matched-pair priority & 0.88456 & 22/2/34 & $-$72.4 [$-$166.3, 11.9] \\
\bottomrule
\end{tabular*}
\end{table}

\paragraph{Results relative to the reference.}
Table~\ref{tab:relation-design-full} reports direct two-method contrasts between each variant and the fixed reference.
The reference wins 41 of 58 tasks against both uniform weighting and the gate threshold of 0.30, with direct Elo intervals above zero.
However, the intervals for removing global-output features, using similarity-only weighting, lowering the gate threshold to 0.10, or removing matched-pair priority include zero.
These results do not establish those individual choices as necessary.
Using 16 anchors beats the 32-anchor reference on 38 of 58 tasks, whereas increasing the cap to 64 is worse overall.
More reference molecules therefore do not guarantee better predictions.
The mean MoleculeACE RMSE difference between 16 and 32 anchors is small (0.88422 versus 0.88496), and no RMSE-difference confidence interval is supplied for this experiment.

\paragraph{Pair prediction and $k$NN transfer with similarity-only weights.}
To separate the benefit of pair prediction from the weighting rule, we compare it with $k$NN residual transfer using similarity-only weights for both methods.
The pair predictor estimates how the query's error differs from each anchor's error; $k$NN transfers the stored anchor errors directly.
Neither side uses relation uncertainty in its anchor weights.
This comparison asks whether learning those error differences adds value under a common weighting rule.

\begin{table}[htbp]
\centering
\footnotesize
\setlength{\tabcolsep}{5pt}
\caption{\textbf{Pair prediction versus neighbor-error transfer.}
RMSE averages 30 MoleculeACE tasks.
W/L and positive direct $\Delta$Elo favor pair prediction across 58 tasks; brackets give its 95\% endpoint-bootstrap interval.}
\label{tab:relation-design-matched}
\begin{tabular*}{\textwidth}{@{\extracolsep{\fill}}lrrrr@{}}
\toprule
Weighting & Pair RMSE & $k$NN RMSE & W/L & $\Delta$Elo [95\% CI] \\
\midrule
Similarity-only & 0.88505 & 0.88899 & 40/18 & +137.7 [47.9, 249.7] \\
\bottomrule
\end{tabular*}
\end{table}

Table~\ref{tab:relation-design-matched} shows that pair prediction wins 40 of 58 tasks, with direct $\Delta$Elo of +137.7 [47.9, 249.7].
Mean MoleculeACE RMSE is 0.88505 for pair prediction and 0.88899 for direct error transfer.
These results support the learned comparison under the matched conditions.
Unlike the full method in the main comparison, pair prediction here does not use relation uncertainty in its weights.
\FloatBarrier

\subsection{Fixed versus similarity-adaptive gating}
\label{app:fixed-gate-results}

\textbf{Purpose.}
The reliability gate in \method{} determines how strongly the molecular-pair prediction changes the global prediction.
Its coefficient increases with the maximum Morgan Tanimoto similarity between the query and its accepted anchors.
Removing the gate applies the full predicted change, but that comparison does not tell us whether the coefficient needs to vary across queries.
Here we ask whether the similarity-based gate improves on using the same fixed coefficient for every query.

\textbf{Matched comparison.}
We reuse the cached predictions from the primary CheMeleon--TabPFN-3 configuration on the 58 structural-generalization tasks.
Global predictions, selected anchors and their weights, relation-TFM predictions, test splits, and seeds remain unchanged; no model is retrained.
We replace only the adaptive gate coefficient with 0.10, 0.20, 0.25, 0.30, 0.40, or 0.50.
Queries without an accepted anchor still return the global prediction in every variant.
We also test a fixed coefficient of 0.2674, the mean adaptive coefficient over active query--seed predictions in this analysis.
This matches the average coefficient without using test labels to choose it.
RMSE is averaged over the 30 MoleculeACE tasks after averaging seeds within each task.
For each adaptive-versus-fixed comparison, direct Elo uses one outcome per task from seed-averaged metrics, with 95\% intervals from resampling complete tasks.

\begin{table}[t]
\centering
\footnotesize
\setlength{\tabcolsep}{5pt}
\caption{\textbf{Adaptive versus fixed reliability-gate coefficients.}
W/L counts adaptive-gate wins/losses on 58 tasks; positive direct $\Delta$Elo favors the adaptive gate.
Brackets give 95\% confidence intervals. Bold marks the lowest mean MoleculeACE RMSE.}
\label{tab:fixed-vs-adaptive-gate}
\begin{tabular}{@{}lccc@{}}
\toprule
Gate & MACE RMSE & Adaptive W/L & Direct $\Delta$Elo [95\% CI] \\
\midrule
Adaptive & \textbf{0.885112} & -- & -- \\
Fixed 0.10 & 0.889325 & 44/14 & $+197.2$ [97.7, 314.1] \\
Fixed 0.20 & 0.886889 & 32/26 & $+35.8$ [$-47.9$, 124.0] \\
Fixed 0.25 & 0.886215 & 27/31 & $-23.8$ [$-110.7$, 60.1] \\
Fixed 0.2674 & 0.886066 & 26/32 & $-35.8$ [$-124.0$, 47.9] \\
Fixed 0.30 & 0.885905 & 24/34 & $-60.1$ [$-151.7$, 23.8] \\
Fixed 0.40 & 0.886378 & 23/35 & $-72.4$ [$-166.3$, 11.9] \\
Fixed 0.50 & 0.888306 & 29/29 & $0.0$ [$-85.0$, 85.0] \\
\bottomrule
\end{tabular}
\end{table}

\textbf{Results.}
Table~\ref{tab:fixed-vs-adaptive-gate} shows that the adaptive gate has the lowest mean MoleculeACE RMSE among the tested coefficients.
The closest fixed alternative is 0.30, with an RMSE difference of approximately 0.00079.
Across all 58 tasks, the adaptive gate wins 44 against fixed 0.10, with a positive direct Elo interval.
However, all Elo intervals for fixed coefficients from 0.20 to 0.50 include zero.
Against the mean-matched coefficient of 0.2674, the adaptive gate wins 26 tasks and loses 32.
Thus, its lower mean regression error does not establish a consistent task-level advantage over moderate fixed coefficients.
The comparison also does not isolate the benefit of fallback, because the same queries fall back to the global prediction in every variant.

\section{Evaluation Beyond the Primary Structural Splits}
\label{app:other-settings}
\label{sec:correction-scope-results}

Our primary benchmark asks whether predictions improve when test molecules are structurally separated from labeled training molecules by scaffold groups and a fingerprint-similarity cutoff.
That result does not establish an advantage for every way in which future molecules can differ from the available data.
The following experiments retain the molecular-pair approach but change the evaluation setting: scaffold-only separation, the official Lo-Hi tasks, DrugOOD domain shifts, and publication-year splits.
We state the prediction problem and comparison for each setting before discussing its results.
Their task populations and metrics differ, so they are reported separately and do not enter the primary leaderboard or its Elo pool.
A complementary analysis of which individual queries improve within the primary benchmark is given in Appendix~\ref{app:cached-support}.

\subsection{Comparison of prediction strategies on scaffold-only splits}
\label{app:scaffold-controls}
\label{app:encoder-protocols}
\label{app:standard-results}

\paragraph{Purpose and setup.}
Our primary protocol removes close train--test analogues in addition to holding out scaffold groups.
Here we ask whether the comparison between pair-based and single-molecule prediction changes when only scaffold separation is imposed.
The scaffold-only partitions keep scaffold groups disjoint but do not apply the additional maximum-similarity filter; their construction is specified in Appendix~\ref{app:primary-splits}.
We retain CheMeleon--TabPFN-3, RDKit2D relation features, and same-fold OOF pairs.
Within this experiment, all compared methods reuse the same saved global features, predictions, OOF outputs, uncertainties, and molecule order.

The alternatives are the gated single-residual model, the single-residual model with matched context dimensions, and the property-delta pair model defined in Appendix~\ref{app:shared-global-results}.
The first two ask whether molecular pairs improve on predicting each molecule's error independently.
The third asks whether predicting error differences is better than predicting property differences while retaining global-prediction information.

\paragraph{Results.}
\method{} has task W/L of 34/24 against the gated single-molecule alternative, 35/23 against the row-matched alternative, and 26/32 against property-delta prediction.
It therefore wins a majority of tasks against the two single-molecule methods, but not against the alternative pair target.
These counts describe how often each method wins, not the magnitude or statistical reliability of its advantage.
They suggest that the choice of pair target can depend on the evaluation setting rather than favoring error differences uniformly.
\FloatBarrier

\subsection{Hit identification and lead optimization}
\label{app:lohi-results}

\paragraph{Purpose and prediction tasks.}
We next ask whether the molecular-pair approach helps on an independently defined benchmark with official partitions, rather than partitions constructed for this paper.
Lo-Hi distinguishes finding promising compounds in a broad candidate set from ranking the properties of closely related compounds~\citep{steshin2023hi}.
Its hit-identification tasks use DRD2, HIV, KDR, and solubility data over 12 released folds and report \ac{PR-AUC}.
Its lead-optimization tasks use DRD2, KCNH2, and KDR over nine released folds and report Spearman correlation within compound clusters.
These measures address different prediction questions and are not combined into one score.

\paragraph{Comparison and results.}
We compare the global CheMeleon--TabPFN-3 predictor with \method{} using RDKit2D relation features and same-fold OOF pairing, keeping Lo-Hi's official data, folds, and evaluators.
Table~\ref{tab:lohi-main} also shows the evaluated fingerprint-based and Chemprop baselines for context.
Values are means across the released folds, with higher values preferred for both metrics.

\begin{table}[!htbp]
\centering
\small
\setlength{\tabcolsep}{4.0pt}
\caption{\textbf{Official Lo-Hi results.} Values are means across released folds; higher is better.}
\label{tab:lohi-main}
\begin{tabular}{@{}lrr@{}}
\toprule
Method & Hit identification & Lead optimization \\
& \acs*{PR-AUC} & Cluster Spearman \\
\midrule
$k$NN with \acs*{ECFP4} & 0.4606 & 0.1579 \\
Gradient boosting with \acs*{ECFP4} & 0.4594 & 0.2140 \\
SVM with \acs*{ECFP4} & 0.4065 & 0.3124 \\
Chemprop & 0.5040 & 0.2172 \\
CheMeleon with TabPFN-3 & 0.5551 & 0.3004 \\
\method{} & \textbf{0.5578} & \textbf{0.3233} \\
\bottomrule
\end{tabular}
\end{table}

\method{} raises mean hit-identification PR-AUC from 0.5551 to 0.5578 and mean lead-optimization Spearman correlation from 0.3004 to 0.3233.
These are small positive mean changes on the official benchmark.
The means alone do not show that every dataset benefits or establish the statistical reliability of the changes.
\FloatBarrier

\subsection{Assay, size, and chronological shifts}
\label{app:external-shifts}

\paragraph{Why evaluate these settings?}
Structural novelty is only one way in which future data can differ from training data.
Molecules may also be measured in different assays, differ in molecular size, or arrive at a later point in time.
We use these settings to test the scope of the observed benefit, not to treat every distribution shift as equivalent to our primary structural split.

\paragraph{DrugOOD.}
DrugOOD supplies six ligand-based IC50 and EC50 tasks covering assay, scaffold, and size shifts~\citep{ji2023drugood}.
We use its official partitions and compare the global predictor and the RDKit2D-relation \method{} configuration alongside the methods in Table~\ref{tab:drugood}.
The table reports mean ROC-AUC across the six tasks, so it is an aggregate over several shift types rather than a result on one homogeneous test population.

\begin{table}[!htbp]
\centering
\small
\setlength{\tabcolsep}{3.4pt}
\caption{\textbf{Official DrugOOD results.} Values are mean ROC-AUC across six tasks; higher is better.}
\label{tab:drugood}
\begin{tabular}{lrrrrrr}
\toprule
Method & MoleOOD & \acs*{ERM} & MixUp & Global \acs*{TFM} & \method{} & GroupDRO \\
\midrule
\acs*{ROC-AUC} & 0.6946 & 0.6739 & 0.6714 & 0.6704 & 0.6687 & 0.6423 \\
\bottomrule
\end{tabular}
\end{table}

MoleOOD has the highest reported mean ROC-AUC, 0.6946~\citep{yang2022learning}.
The global predictor scores 0.6704 and \method{} scores 0.6687.
Thus, adding the molecular-pair stage does not improve the mean performance in this evaluation, despite the gains under our primary protocol.

\paragraph{Publication-year splits.}
Two further ChEMBL tasks separate EC50 and IC50 measurements by publication year.
Unlike the scaffold-plus-similarity rule, this defines the training--test distinction through chronology rather than a chosen degree of structural separation.
We compare the same global and pair-based predictors and average their five paired evaluation seeds within each task.
The global and \method{} ROC-AUC values are 0.868307 and 0.868315 for EC50, and 0.706821 and 0.703370 for IC50.
EC50 is essentially unchanged, whereas IC50 decreases.
Although the seed-averaged scores give one task win and one loss, the sizes of those changes are very different.
Together with DrugOOD, these results limit our conclusion to the settings actually supported by the evidence: molecular comparisons do not improve predictions under every type of molecular distribution shift.
\FloatBarrier

\section{Prediction Changes and Serving Costs}
\label{app:diagnostics}

Task-level averages do not show which individual molecules benefit or how much it costs to use an already prepared predictor.
The following analyses answer those questions separately.
They retain the distinction between a molecule-level prediction, a task-level metric, and an end-to-end workflow cost rather than interpreting them as interchangeable summaries.

\subsection{Which queries improve, and when is the global prediction retained?}
\label{app:cached-support}
\label{app:chemical-support}

\paragraph{Purpose: similarity to the reference molecules.}
\method{} uses labeled training molecules as anchors when adjusting a query's global prediction.
We ask how often this helps when at least one selected anchor is structurally similar to the query, and whether improvements also occur when even the closest selected anchor is less similar.
Here similarity means Morgan-fingerprint Tanimoto similarity between the query and a selected labeled reference molecule; it does not mean similarity between predicted property values.
For each query, we use the highest similarity among its accepted anchors and group predictions into the intervals shown in Table~\ref{tab:cached-similarity-effects}.
A query with no accepted anchor is placed in a separate group and receives the unchanged global prediction.

\paragraph{Data and comparison.}
This is an analysis of saved predictions from the primary experiment, not a new split or an additional model run.
The records cover 286 task--seed runs across 58 tasks: 231 regression runs and 55 classification runs.
They contain 42,026 regression and 13,707 classification predictions, totaling 55,733.
The same molecule can appear under multiple evaluation seeds, so these are query--seed observations rather than unique molecules.
For each observation, we compare \method{} with its corresponding global prediction on the same query.
Improvement means lower absolute error for regression or lower Brier loss for classification; Brier loss is the squared difference between the predicted positive-class probability and the binary label.
Each query--seed observation receives equal weight in this descriptive analysis, unlike the equal-task weighting used for Elo.

\begin{table}[t]
\centering
\small
\setlength{\tabcolsep}{5pt}
\caption{\textbf{Correction outcomes by anchor similarity.} $N$ counts query--seed predictions. Improved/worsened denotes lower/higher absolute error (regression) or Brier loss (classification).}
\label{tab:cached-similarity-effects}
\begin{tabular}{@{}lrrrrrr@{}}
\toprule
& \multicolumn{3}{c}{Regression} & \multicolumn{3}{c}{Classification} \\
\cmidrule(lr){2-4}\cmidrule(l){5-7}
Similarity & $N$ & Improved (\%) & Worsened (\%) & $N$ & Improved (\%) & Worsened (\%) \\
\midrule
$[0.20,0.30)$ & 6,824 & 57.1 & 41.9 & 2,494 & 64.8 & 33.0 \\
$[0.30,0.40)$ & 12,415 & 57.2 & 42.8 & 3,624 & 67.4 & 32.6 \\
$[0.40,0.50)$ & 10,596 & 56.5 & 43.5 & 3,757 & 70.0 & 30.0 \\
$[0.50,0.60)$ & 11,298 & 54.8 & 45.2 & 3,071 & 74.0 & 26.0 \\
$=0.6$ & 419 & 53.2 & 46.8 & 104 & 71.2 & 28.8 \\
No anchor & 474 & 0.0 & 0.0 & 657 & 0.0 & 0.0 \\
\bottomrule
\end{tabular}
\end{table}

\paragraph{How often do predictions improve?}
Across similarity intervals from 0.20 to below 0.60, \method{} improves 54.8--57.2\% of regression predictions and 64.8--74.0\% of classification predictions.
A substantial minority worsen.
The regression improvement fraction does not consistently increase with structural similarity, so a closer reference does not by itself imply that a query will benefit.
These fractions describe the frequency of gains and harms, not their magnitude.
They also compare different groups of molecules rather than varying similarity for the same molecule, so they do not isolate a causal effect of similarity.

We do not average absolute-error magnitudes across regression tasks with different target units.
For classification, the mean Brier-loss reductions across the five nonempty similarity intervals are 0.000939, 0.001959, 0.000332, 0.000116, and $-0.008174$.
In the last exported interval, 71.2\% of 104 predictions improve, yet mean Brier loss increases.
This illustrates why a high improvement frequency can coexist with larger losses on a smaller number of queries.

\paragraph{How often is the pair model actually used?}
The method can fall back to the global predictor when no anchor passes the acceptance rule.
We therefore check whether the aggregate gain mostly comes from changing predictions or from frequently leaving them untouched.
Table~\ref{tab:cached-gate-fallback} weights each run summary by its test-set size, matching the query--seed weighting above.
Its mean anchor counts include queries with no accepted anchor.

\begin{table}[t]
\centering
\small
\setlength{\tabcolsep}{6pt}
\caption{\textbf{Anchor availability and correction activity.} Fallback means no accepted anchor; active correction means a nonzero change. Percentages and means weight query--seed predictions equally and include fallback cases.}
\label{tab:cached-gate-fallback}
\begin{tabular}{@{}lrrr@{}}
\toprule
Diagnostic & Regression & Classification & All \\
\midrule
Endpoints & 47 & 11 & 58 \\
Task--seed runs & 231 & 55 & 286 \\
Query--seed predictions & 42,026 & 13,707 & 55,733 \\
No-anchor predictions & 474 & 657 & 1,131 \\
No-anchor fallback (\%) & 1.13 & 4.79 & 2.03 \\
Active correction (\%) & 98.71 & 94.81 & 97.75 \\
Mean anchor count & 26.66 & 22.66 & 25.68 \\
\bottomrule
\end{tabular}
\end{table}

There are 474 no-anchor regression predictions and 657 no-anchor classification predictions, totaling 1,131 (2.03\%).
All have unchanged loss.
In total, 1,253 predictions are unchanged: the 1,131 fallback cases and another 122 cases with an available anchor.
An accepted anchor therefore does not guarantee a nonzero adjustment, for example at the gate boundary.
Overall, 97.75\% of predictions receive a nonzero adjustment.
The overall gains thus arise in a setting where most predictions are actively changed, with both beneficial and harmful changes, rather than one dominated by fallback.

As a separate summary, we average run-level percentages within each task and then give tasks equal weight.
This gives a 3.22\% no-anchor rate and a 96.46\% nonzero-adjustment rate, compared with 2.03\% and 97.75\% when observations receive equal weight.
The difference arises because large test sets contribute more to the observation-weighted summary; neither percentage should be substituted for the other without changing its interpretation.

\subsection{Predictions on activity-cliff molecules}
\label{app:activity-cliff-results}

\paragraph{Purpose and setup.}
A structurally similar molecule need not have a similar measured property.
Activity cliffs describe cases where similar molecular structures have substantially different properties, making them a relevant check for a method that uses molecular references.
We use the activity-cliff annotations supplied with MoleculeACE to select the corresponding test molecules under the primary structural-generalization protocol.
We compare the saved predictions of \method{} and its global baseline on those molecules; this does not train a separate cliff detector or change anchor selection.

\paragraph{Results.}
After averaging seeds within each task and weighting the 30 MoleculeACE tasks equally, mean cliff RMSE is 0.9565 for \method{} and 0.9706 for the global predictor.
The lower mean is descriptive: no paired confidence interval is reported for this subset analysis.
It does not establish a consistent gain on every task, or show that \method{} explicitly recognizes cliffs or resolves abrupt property changes between a query and its anchors.

\subsection{Setup, label updates, and repeated query costs}
\label{app:cost-results}

\paragraph{Purpose and comparison.}
The main performance--cost figure includes the work needed to select and prepare a predictor for a task.
Once that work has been done, a different question matters: how much time is needed to update the task or predict another batch of molecules?
We compare the global predictor, the primary CheMeleon--TabPFN-3 \method{} configuration, and a fitted MiniMol head in a separate six-endpoint timing study.
Table~\ref{tab:latency} reports median seconds across those endpoints, not the mean workflow times across the 58 primary tasks used in Figure~\ref{fig:performance-cost}.

The measured scenarios distinguish cold end-to-end execution, warm task adaptation, an update with 64 labels, and warm query batches of 1, 128, or 10,000 molecules.
The query-batch measurements examine serving an already prepared predictor rather than repeating hyperparameter search for every batch.
These timings are kept separate from the main Pareto analysis because both the endpoint population and the work being measured differ.

\begin{table}[!htbp]
\centering
\small
\setlength{\tabcolsep}{3.2pt}
\caption{\textbf{Setup and query runtime.} Values are median seconds across six endpoints.}
\label{tab:latency}
\begin{tabular}{lrrr}
\toprule
Scenario & Global & \method{} & MiniMol head \\
\midrule
Cold end to end & 5.17 & 46.11 & 13.91 \\
Warm task adaptation & 0.17 & 13.41 & 8.60 \\
Update with 64 labels & 0.15 & 4.49 & 2.09 \\
Warm query with 1 molecule & 1.10 & 1.64 & 0.002 \\
Warm query with 128 molecules & 1.14 & 3.91 & 0.003 \\
Warm query with 10,000 molecules & 7.16 & 228.95 & 0.004 \\
\bottomrule
\end{tabular}
\end{table}

\paragraph{Results.}
Warm task adaptation takes 13.41 seconds for \method{}, compared with 8.60 seconds for the MiniMol head and 0.17 seconds for the global predictor.
Both comparators are faster than \method{} in every reported scenario.
For 10,000 warm queries, \method{} takes 228.95 seconds, compared with 7.16 seconds for the global predictor.
The pair stage therefore adds a recurring inference cost, not only a one-time setup cost.
This does not contradict its favorable comparison with task-specific hyperparameter search: preparing a method and repeatedly serving it are different cost comparisons.
The main figure supports the end-to-end performance--cost tradeoff, while this study makes the additional setup and serving costs explicit.
\FloatBarrier

\section{Reproducibility and Result Provenance}
\label{app:reproducibility}

\subsection{Software and hardware}
\label{app:software}

\paragraph{Environment.}
The reference environments use Python 3.11, RDKit 2026.3.4, XGBoost 3.1.3, LightGBM 4.6.0, CatBoost 1.2.8, Optuna 4.9.0, PyTorch 2.13.0, TabPFN 8.2.0, TabICL 2.1.1, Chemprop 2.3.0, MordredCommunity 2.0.7, Polaris 0.13.0, and DataSAIL 1.4.0.
The environments use scikit-learn 1.6 to 1.8.
Exact conda specifications cover the main, MiniMol, molecular-model, DataSAIL, and TDC environments with their incompatible upstream dependencies.

\paragraph{Hardware and scheduler.}
Primary \ac{GPU} evaluations use one NVIDIA H200 per worker.
Matrix multiplication uses high precision with TF32 enabled.
Runs use fixed seeds when strict kernel determinism is unavailable, so identical settings do not guarantee bitwise-identical predictions.
\ac{GPU} feature and model work is pinned to the \ac{GPU} \ac{NUMA} node, while \ac{CPU} tree workers use the other \ac{NUMA} node.
The scheduler records \ac{GPU} utilization, memory, process state, outcome, and timing.
Content-addressed records let interrupted manifests resume without repeating validated outputs.

\subsection{Data integrity and execution}
\label{app:integrity}

\paragraph{Artifact integrity.}
Each job and cache identity includes dataset revisions, benchmark checksums, checkpoint hashes, row-order hashes, preprocessing rules, method configuration, and environment versions.
Prediction records store the task, split, method, representation, seed, metric, metric direction, timing, and applicability status.
Before fitting Elo, aggregation rejects duplicate task-method-seed rows and checks that the compared methods cover the required paired observations.

\paragraph{Split integrity.}
The outer training partition supplies every learned transformation, cross-fitting fold, anchor table, and hyperparameter choice.
Paired methods use the same retained row identifiers.
For every endpoint and seed, the structural-generalization audit records test-set size, scaffold overlap, maximum training similarity, and exclusion reasons.
The DataSAIL fingerprint study also archives the \ac{ECFP4} and \ac{MACCS} similarity matrices together with the validated partition assignments.

\end{document}